\documentclass[12pt, letterpaper]{article}
\usepackage[utf8]{inputenc}
\usepackage[margin=1in]{geometry}
\usepackage{booktabs}
\usepackage{tabu}
\usepackage[labelfont=bf, skip=5pt, font=small]{caption}
\usepackage{subcaption}
\usepackage{graphicx}
\usepackage{fancyhdr}
\usepackage{authblk}
\usepackage{algorithm}
\usepackage{algorithmicx}
\usepackage{algpseudocode}
\usepackage{multicol,lipsum}
\usepackage{float}
\usepackage{placeins}
\usepackage{tabularx}
\usepackage{tikz}
\usetikzlibrary{shapes.geometric, arrows, decorations.pathreplacing,calligraphy}
\usepackage{lscape}
\usepackage{multirow}
\usepackage[margin=1in]{geometry}
\usepackage{booktabs}
\usepackage{tabu}
\usepackage[labelfont=bf, skip=5pt, font=small]{caption}
\usepackage{graphicx}
\usepackage{fancyhdr}
\usepackage{biblatex}
\usepackage{rotating}
\usepackage[colorlinks=true, allcolors=black]{hyperref}
\usepackage{pdfpages}
\usepackage{nomencl}
\usepackage{xcolor}
\usepackage{longtable}
\DeclareUnicodeCharacter{2212}{-}
\usepackage{amsmath}

\author[a]{Ruturaj Reddy\footnote{1132200213@mitwpu.edu.in}}
\author[a*]{Anand J Kulkarni\thanks{anand.j.kulkarni@mitwpu.edu.in}}
\author[b]{Ganesh Krishnasamy\footnote{ganesh.krishnasamy@monash.edu}}
\author[a]{\\ Apoorva S Shastri\footnote{apoorvasapre@gmail.com}}
\author[c]{Amir H. Gandomi\footnote{amirhossein.gandomi@uts.edu.au}}
\affil[a]{\small Institute of Artificial Intelligence, MIT World Peace University, 124 Paud Road, Kothrud, Pune 411038, MH, India}
\affil[b]{School of Information Technology, Monash University, Jalan Lagoon Selatan, Bandar Sunway, 47500 Subang Jaya, Selangor, Malaysia}
\affil[c]{Faculty of Engineering \& Information Technology, University of Technology Sydney, Ultimo, Australia\vspace{-3.5em}}
\date{}

\providecommand{\keywords}[1]
{
  \small	
  \textbf{\textit{Keywords---}} #1
}

\title{\vspace{-1.8cm}LAB: A Leader-Advocate-Believer Based Optimization Algorithm\vspace{-0.3cm}}
\makenomenclature

    \makeatletter
\def\@fnsymbol#1{\ensuremath{\ifcase#1\or \dagger\or \ddagger\or
   \mathsection\or \mathparagraph\or \|\or **\or \dagger\dagger
   \or \ddagger\ddagger \else\@ctrerr\fi}}

\begin{document}
\maketitle
\vspace{-0.2cm}
\rule{\textwidth}{0.5pt}
\begin{abstract}
    \noindent This manuscript introduces a new socio-inspired metaheuristic technique referred to as Leader-Advocate-Believer based optimization algorithm (LAB) for engineering and global optimization problems. The proposed algorithm is inspired by the AI-based competitive behaviour exhibited by the individuals in a group while simultaneously improving themselves and establishing a role (Leader, Advocate, Believer). LAB performance in computational time and function evaluations are benchmarked using other metaheuristic algorithms. Besides benchmark problems, the LAB algorithm was applied for solving challenging engineering problems, including abrasive water jet machining, electric discharge machining, micro-machining processes, and process parameter optimization for turning titanium alloy in a minimum quantity lubrication environment. The results were superior to the other algorithms compared such as Firefly Algorithm, Variations of Co-hort Intelligence, Genetic Algorithm, Simulated Annealing, Particle Swarm Optimisation, and Multi-Cohort Intelligence. The results from this study highlighted that the LAB outperforms the other algorithms in terms of function evaluations and computational time. The prominent features of the LAB algorithm along with its limitations are also discussed.
    \vspace{-0.1cm}
\end{abstract}
\rule{\textwidth}{0.5pt}


\section{Introduction}
Optimization is a way of finding the best solutions to most of the problems encountered in real life. On a regular basis we encounter problems where we try to minimize efforts and maximize outcomes \cite{ABUALIGAH2021113609} on an action, may it be driving to work on a specific road at a specific time to minimise the time required to reach destination or decrease speed to increase mileage. Other than day-to-day implementations optimization is used on a larger scale too, such as manufacturing of cars in order to minimise wind resistance and maximise speed and handling or designing products in such a way to minimise material cost and maximise the quality and profits, etc. A variety of optimization methods inspired by nature have been developed to solve these problems\cite{inbook, optiintro}. These algorithms can be classified into four major categories: biology-inspired/bio-inspired, swarm intelligence, socio-inspired and physics/chemistry-based\cite{Fister2013ABR}.

\begin{enumerate}
\item {\textit{Bio-inspired Intelligence Techniques}: These algorithms are inspired by biological evolution and species. The most well-known and widely used bio-inspired algorithm is the  Genetic Algorithm (GA)\cite{holland1992adaptation}. It is based on the Darwinian theory of survival of the fittest\cite{genetic}. The algorithm relies on three important factors, mutation, crossover and selection to approach better quality solutions. Other examples are Covariance Matrix Adaptation Evolution (CMA-ES)\cite{10.1162/evco.2007.15.1.1} based on basic genetic rules, Backtracking Search Algorithm (BSA)\cite{CIVICIOGLU20138121}, Evolutionary Strategies (ES)\cite{Beyer2004EvolutionS}, Evolutionary Programming\cite{Fogel:2011}, Differential Evolution (DE)\cite{DE,Qin2005SelfadaptiveDE}, inspired by biological evolutionary strategies such as reproduction, mutation, recombination and selection\cite{michalewicz1996evolutionary}. A variety of optimization methods inspired by nature have been developed to achieve better solutions than current methods. BSA is amongst the recently proposed algorithms, which generates a trial individual using basic genetic operators (selection, mutation and crossover). Evolutionary programming is one of the first genetic algorithms developed; however evolutionary programming differs from standard GA as the focus is on the behavior of individuals, thus no crossover is used. DE is a population-based stochastic function minimizer based on iterating population towards a quality goal. JDE\cite{JDE}, JADE\cite{5208221} and SADE\cite{Qin2010} are recent versions of DE.}

\item {\textit{Swarm Based Intelligence Techniques}: Swarm intelligence (SI) refers to a subset of bio-inspired techniques. The individuals in the swarm collectively organize themselves to achieve a common goal \cite{IGLESIAS2020273}. Particle Swarm Optimization (PSO) is one of the popular swarm intelligence methods \cite{PSO}. It is inspired from the schooling of fish. In PSO, it starts with random initialisation of population and moves to search optima while updating generations. PSO uses parameters of social and individual behaviors as opposed to evolution operators used in GA. CLPSO\cite{CLPSO} and PSO2011\cite{particle} are the updated versions of the standard PSO. Other examples of swarm-intelligence include Cuckoo Search (CS)\cite{cuckoo}, Bat Algorithm (BA)\cite{bat}, Ant Colony Optimization (ACO)\cite{antcolony}, Firefly Algorithm (FA)\cite{yang2010nature,2013}, Artificial Bee Colony (ABC)\cite{KARABOGA2009108}. ACO is based on the excretion of pheromones by ants which helps guide the way for other ants in the system. In FA, all the fireflies are unisexual and are attracted towards higher intensity(brightness) or the flash signals produced while moving towards better search space and decreasing distance between them. ABC is based on the behaviour of honey bees when discovering food sources.}

\item {\textit{Physics Based Intelligence Techniques}: 
Some algorithms are nature-inspired but are based on principles of physicssuch as laws of gravitation by Newton. Existing physics-based algorithms are \cite{Natureinspired}, Colliding Bodies Optimisation (CBO)\cite{CBO} formulated based on Newton's law of motion, Gravitational Search Algorithm (GSA)\cite{GSA}, Central Force Optimisation (CFO) \cite{CFO}, Space Gravitation Optimisation (SGO) \cite{SGO} and  \cite{GIO} formulated based on Newton’s gravitational force, Big Bang–Big Crunch search (BB–BC) \cite{BBBC},  Galaxy-based Search Algorithm \cite{GBSA} and Artificial Physics-based Optimisation (APO) \cite{APO} formulated based on celestial mechanics and astronomy, Ray Optimisation (RO) \cite{RO} is based on optics, Harmony Search Algorithm (HSA) \cite{HSA} formulated based on acoustics, Simulated Annealing (SA) algorithm is based on thermodynamics principle \cite{SA}.}

\item {\textit{Socio-inspired Intelligence Techniques}:
The Cultural/Social Algorithm is a subset of evolutionary\textcolor{blue}{-}based intelligence. In a society, humans learn from one another by following them which eventually helps them evolve and achieve their goals together\cite{socioevo}. Based on these motives, many researchers began to develop social/socio-inspired algorithms, such as Society and Civilization Optimization Algorithm (SCO)\cite{SCO}, Imperialist Competitive Algorithm (ICA) \cite{ICA}, League Championship Algorithm (LCA) \cite{LCA}, Cultural Evolution Algorithm (CEA) \cite{CEA}, Cohort Intelligence (CI) \cite{CI}, Social learning Optimization (SLO) \cite{SLO}, Social Group Optimization (SGO) \cite{SGO2} and Ideology Algorithm (IA), etc.}
\end{enumerate}
In this manuscript, the work is based on the competitive behaviour of individuals within a group in a competitive environment that has existed in human society for ages, may it be at an academic level or corporate level. The ultimate goal of an individual in a group is to establish his/her position also, known as rank, by competing with other individuals within the group while moving towards promising directions.\\
This manuscript introduces a novel socio-inspired optimization algorithm referred to as LAB: A Leader-Advocate-Believer-based optimization algorithm. The society individuals are divided into groups and are categorised into certain roles. These groups and roles help by guiding a way for the individuals to achieve their goals by competing with the individuals within the corresponding group while moving towards a promising search space. The LAB is motivated by this competitive trait of individuals in a group. Every group leader moves in a certain direction that motivates individuals to compete with it in order to lead the group towards a more promising search space. Not only does every individual compete with its associated leader but it also competes with the individuals within its group with a goal to improve and promote to a higher rank. However, the short-term goal of an individual is to reach as close as possible to its local leader. Furthermore, every local group leader always desires to be the global best leader; thus, it competes with other group leaders to become the global leader while competing with the rising individuals within its associated group. This competitive behaviour of an individual increases its chances of improving and climbing up in the group while moving towards promising search spaces is modelled here. This mechanism enabled LAB to solve several benchmark problems as well as real world problems from manufacturing domain. The performance of the LAB algorithm was better in terms of objective function as well as computational cost as compared to the existing algorithms.

The rest of the manuscript is structured as follows: Section 2 describes the methodology of the LAB algorithm with its flowchart(fig.\ref{fig:flowchart}). Section 3 discusses the benchmark test problems, real-world machining problems.  as well as individual problem formulations and a description of the processes. The performance analysis and comparison of algorithms are discussed in Section 4. In Section 5 concluding remarks and future directions are provided.

\section{LAB Algorithm}

\nomenclature{\(\textit{\textbf{P}}\)}{population of society}
\nomenclature{\(n\)}{number of individuals in each group}
\nomenclature{\(\textit{\textbf{G}}\)}{total number of groups}
\nomenclature{\(f(\textit{\textbf{X}})\)}{objective function}
\nomenclature{\(\psi^l\)}{lower bound}
\nomenclature{\(\psi^u\)}{upper bound}
\nomenclature{\(p\)}{individual}
\nomenclature{\(p_{L_{g}}\)}{leader for the $g^{th}$ group}
\nomenclature{\(p_{L_{1}}^{*}\)}{global best leader}
\nomenclature{\(p_{B_i}^{p_{L_{g}}}\)}{$i^{th}$ believer associated to the leader of the $g^{th}$ group}
\nomenclature{\(p_{A}^{p_{L_{g}}}\)}{advocate associated to the leader of the $g^{th}$ group}
\nomenclature{\(\textit{\textbf{p}}_{B}^{p_{L}}\)}{set of believers}
\nomenclature{\(\textit{\textbf{P}}_L\)}{set of leaders}
\nomenclature{\(\textit{\textbf{P}}_A\)}{set of advocates}
\nomenclature{\(R_a\)}{surface roughness}
\nomenclature{\(kerf\)}{taper angle}
\nomenclature{\(MRR\)}{material remove rate}
\nomenclature{\(REWR\)}{electrode wear rate}
\nomenclature{\(f_b\)}{flank wear}
\nomenclature{\(M_t\)}{machining time}
\nomenclature{\(B_h\)}{burr height}
\nomenclature{\(B_t\)}{burr thickness}
\nomenclature{\(\phi\)}{approach angle}
\nomenclature{\(V_c\)}{cutting speed}
\nomenclature{\(f\)}{feed rate}
\nomenclature{\(F_c\)}{tangential force}
\nomenclature{\(V_{Bmax}\)}{tool wear}
\nomenclature{\(L\)}{tool-chip contact length}
\nomenclature{\(w\)}{weight}

\printnomenclature

In the proposed LAB algorithm, every individual in a group competes with every other individual within the group to become the best individual. The position of the individual depends on the fitness/objective function value. The individual with the best fitness value in a group is assigned as the local group leader for the corresponding group and the individuals within the associated group will follow its direction. The second best individual is assigned as the advocate to the leader and the remaining individuals in the group are referred to as believers. The local leader also competes with all the other local leaders from corresponding groups to become the global best leader. All the other local leaders follow the direction of the global best leader while competing with one another. The local rankings motivate the group leader explore promising search spaces and the global rankings forces all the leaders to explore promising search spaces in order to remain the global best leader, while competing with other leaders. This makes every individual within the group compete with one another, thus motivating it to grow and search for better solutions.

\begin{figure}[H]
    \centering
    \includegraphics[width=0.84\linewidth]{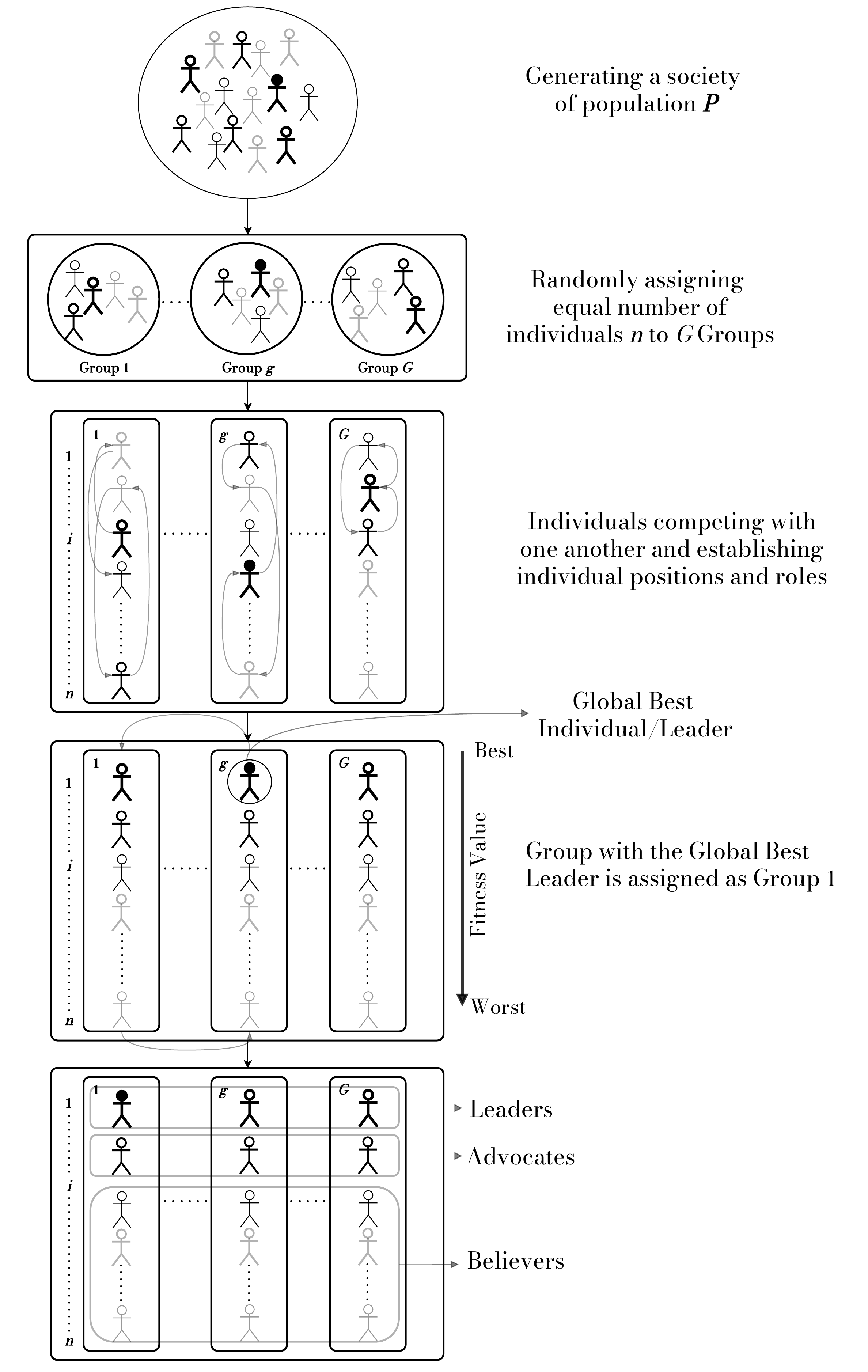}
    \caption{Visual Abstract of the LAB Algorithm}
    \label{fig:visual}
\end{figure}

Consider a general optimization problem as follows:

\begin{center}
Minimize \;\;\;\;\;\;\;\;$f(\textit{\textbf{X}}) = f(x_1,..,x_i,...x_N)$

s.t. \;\;\;\;\;\; $\psi_{i}^l \leq x_i \leq \psi_{i}^u, \; \; \;\;i=1,...,N$    
\end{center}

\begin{figure}[!htb]
\centering
\includegraphics[width=0.50\linewidth]{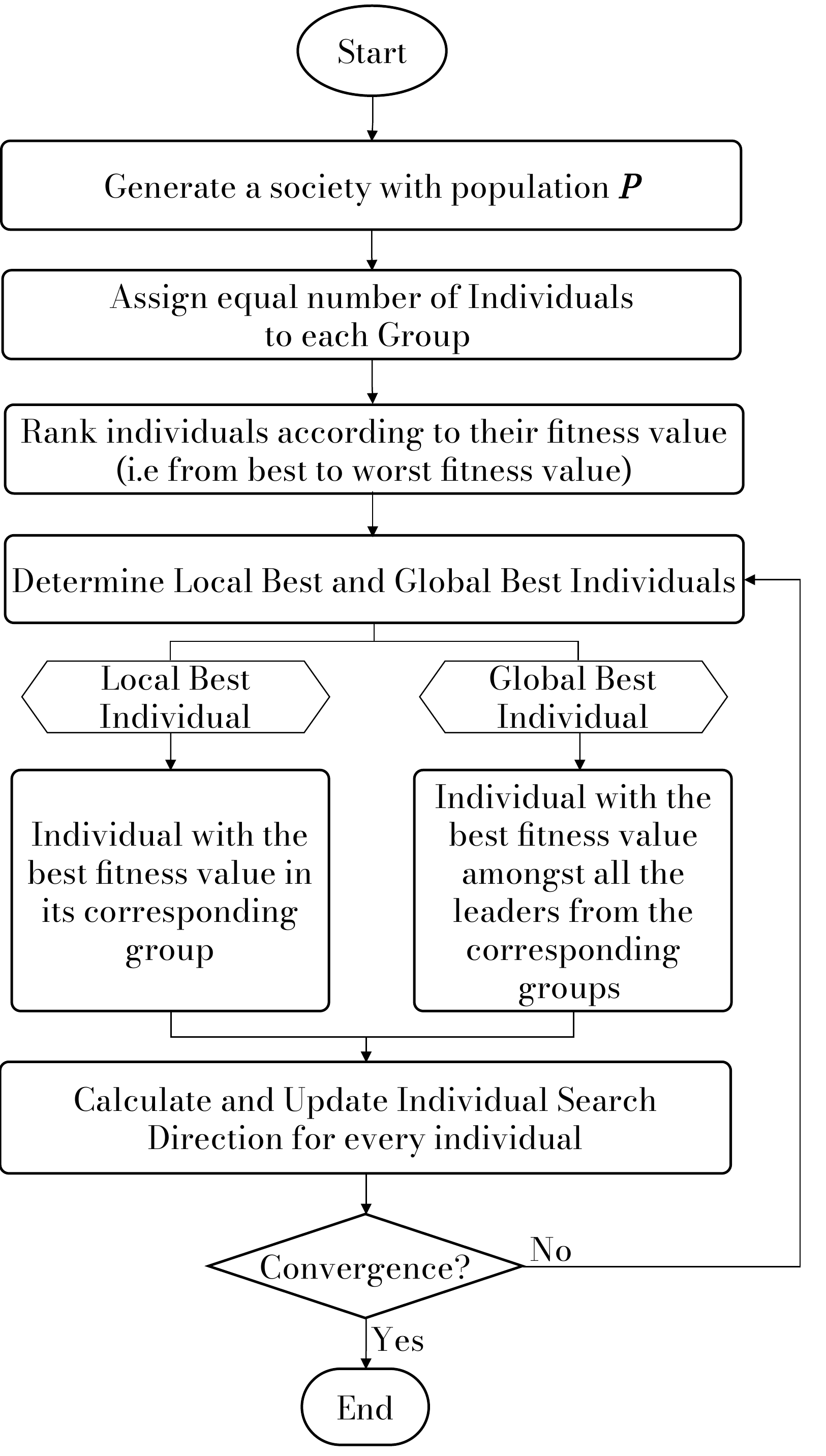}
\caption{LAB Algorithm flowchart}
\label{fig:flowchart}
\end{figure}

The procedure begins with generating a society of population \textit{\textbf{P}} with individuals $p = 1,.....P$ randomly within its associated search space [$\psi_i^{l}$, $\psi_i^{u}$] and associated objective functions are evaluated. Rest of the steps in the algorithm are explained below along with a flowchart (refer to fig.\ref{fig:flowchart})

\textbf{Step 1 (Assigning Groups and establishing Roles) :}
Every group is assigned with an equal number of randomly selected individuals. Each group consists of $n$ number of individuals,

\begin{center}
    where $n = \dfrac{total \;\;number \;\;of \;\; individuals (P)}{total \;number\; of\; groups(G)}$
\end{center}

thus making sure equal number of individuals in each group. After being assigned a group, individuals are locally ranked according to the fitness of their solution (objective function value) and arranged accordingly, i.e. individual with the best fitness quality referred to as Leader ($p_{L}$) followed by second best individual referred to as Advocate ($p_{A}^{p_{L}}$) and remaining individuals ($n-2$, since the first two have been assigned) with worse fitness quality ($p_{B_i}^{p_{L}}$) referred to as Believer. Local best individuals/Leaders from corresponding groups compete with one another. The leader with the best fitness solution is assigned as Global Best Leader ($p_{L}^*$) and the associated group is assigned as Group 1, all the other leaders from the corresponding group follow it's direction. A visual representation for a society of groups with an equal number of individuals is shown below in fig.\ref{fig:sets}.

\begin{figure}[H]
    \tikzstyle{process} = [rectangle, minimum width=3.7cm, minimum height=1cm, text centered, draw=black, fill=white!20]
\tikzstyle{pr} = [rectangle, minimum width=1.2cm, minimum height=0.2cm, text centered, draw=white, fill=white!20]

\begin{tikzpicture}[node distance=1cm]
\node (start) [process, xshift=0cm] {Leader $p_{L_{1}}^{*} $};
\node (in1) [process, below of=start] {Advocate $p_{A}^{p_{L_1}}$};
\node (pro1) [process, below of=in1] {Believer $p_{B_1}^{p_{L_1}}$};
\node (dec1) [pr, below of=pro1] {\textbf{:}};
\node (pro2a) [pr, below of=dec1] {\textbf{:}};
\node (stop221) [pr, rotate=90, below of=pro2a,xshift=0.5cm,yshift=1.0cm] {\textbf{.........}};
\node (pro2a1) [process, below of=pro2a] {Believer $p_{B_n}^{p_{L_1}}$};

\node (pro2b1) [pr, right of=start, xshift=1.5cm] {\textbf{......}};
\node (out11) [pr, below of=pro2b1] {\textbf{......}};
\node (stop1) [pr, below of=out11] {\textbf{......}};
\node (stop11) [pr, below of=stop1] {\textbf{......}};
\node (stop21) [pr, below of=stop11] {\textbf{......}};
\node (stop212) [pr, below of=stop21] {\textbf{......}};

\node (pro2b) [process, right of=start, xshift=4cm] {Leader $p_{L_{g}}$};
\node (out1) [process, below of=pro2b] {Advocate $p_{A}^{p_{L_g}}$};
\node (stop) [process, below of=out1] {Believer $p_{B_1}^{p_{L_g}}$};
\node (stop1) [pr, below of=stop] {\textbf{:}};
\node (stop12) [pr, below of=stop1] {\textbf{:}};
\node (stop221) [pr, rotate=90, below of=stop12,xshift=0.5cm,yshift=1.0cm] {\textbf{.........}};
\node (stop2) [process, below of=stop12] {Believer $p_{B_n}^{p_{L_g}}$};

\node (pro2b2) [pr, right of=pro2b, xshift=1.5cm] {\textbf{......}};
\node (out12) [pr, below of=pro2b2] {\textbf{......}};
\node (stop2) [pr, below of=out12] {\textbf{......}};
\node (stop12) [pr, below of=stop2] {\textbf{......}};
\node (stop22) [pr, below of=stop12] {\textbf{......}};
\node (stop221) [pr, below of=stop22] {\textbf{......}};

\node (gp3) [process, right of=pro2b, xshift=4cm] {Leader $p_{L_{G}}$};
\node (gp31) [process, below of=gp3] {Advocate $p_{A}^{p_{L_G}}$};
\node (gp32) [process, below of=gp31] {Believer $p_{B_1}^{p_{L_G}}$};
\node (gp33) [pr, below of=gp32] {\textbf{:}};
\node (gp34) [pr, below of=gp33] {\textbf{:}};
\node (stop221) [pr, rotate=90, below of=gp33,xshift=-0.5cm,yshift=1.0cm] {\textbf{.........}};
\node (gp35) [process, below of=gp34] {Believer $p_{B_n}^{p_{L_G}}$};

\draw [decorate,decoration = {brace},xshift=-2.2cm,yshift=-3.5cm] (0,3) --  (0,3.9)
node[xshift=-1.0cm,yshift=-0.47cm,black]{Leaders};

\draw [decorate,decoration = {brace},xshift=-2.2cm,yshift=-3.5cm] (0,2) --  (0,2.9)
node[xshift=-1.2cm,yshift=-0.40cm,black]{Advocates};

\draw [decorate,decoration = {brace},xshift=-2.2cm,yshift=-3.5cm] (0,-2) --  (0,1.9)
node[xshift=-1.2cm,yshift=-1.85cm,black]{Believers};

\draw [decorate,decoration = {brace},xshift=-2.2cm,yshift=-4.5cm] (14,-1.3) --  (0.38,-1.3)
node[xshift=6.4cm,yshift=-0.57cm,black]{Society with population \textit{\textbf{P}}};

\draw [decorate,decoration = {brace},xshift=-2.2cm,yshift=-3.5cm] (0.3,4.3) --  (4,4.3)
node[xshift=-1.8cm,yshift=0.57cm,black]{Group 1};

\draw [decorate,decoration = {brace},xshift=-1.8cm,yshift=-3.5cm] (4.8,4.3) --  (8.7,4.3)
node[xshift=-1.8cm,yshift=0.57cm,black]{Group $g$};

\draw [decorate,decoration = {brace},xshift=-1.2cm,yshift=-3.5cm] (9.2,4.3) --  (13.1,4.3)
node[xshift=-2.0cm,yshift=0.57cm,black]{Group G};

\end{tikzpicture}
    \caption{Visual representation of Groups and Roles of Individuals for a Society}
    \label{fig:sets}
\end{figure}
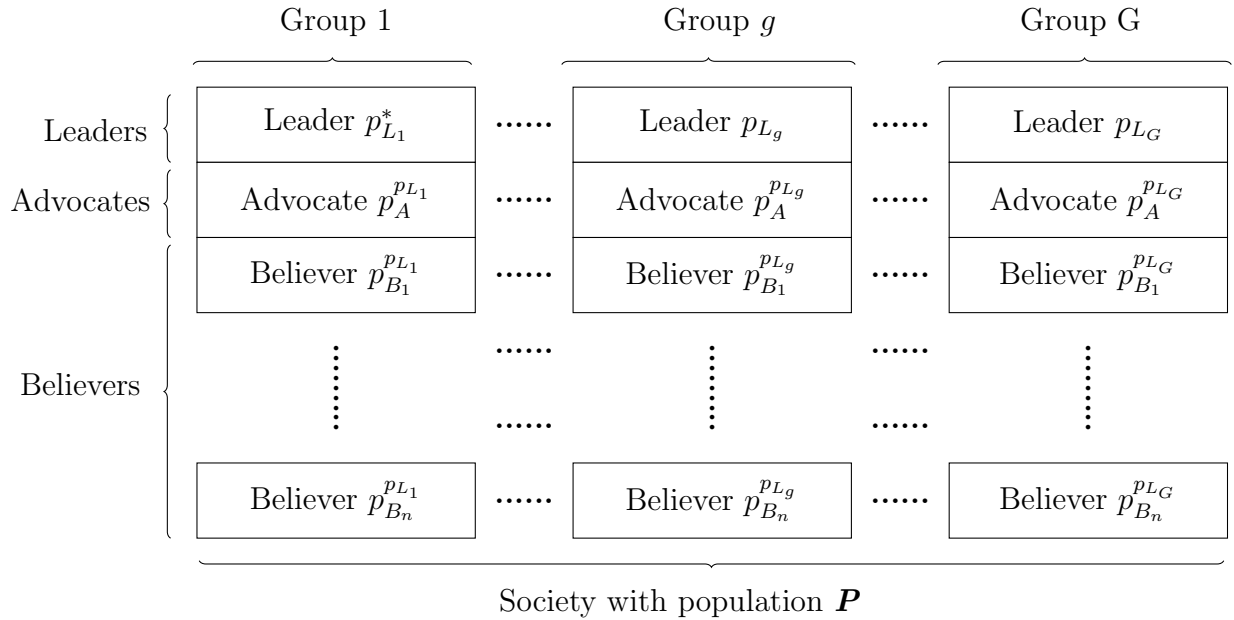

\textbf{Step 2 (Individual Search Direction) :}
With every iteration a new search direction for each individual is calculated, the formulation of the search direction varies as per individual's role as shown below:
\begin{itemize}
\item Leader : The search direction of every leader $p_L \in \textit{\textbf{P}}_L$ is influenced by the global leader $p_{L}^* \in \textbf{\textbf{P}}_L$, the corresponding advocate individual ${p}_{A}^{p_{L}}$, every associated believer $p_{B_{i}}^{p_{L}} \in {\textit{\textbf{p}}}_{B}^{p_{L}}$ and the associated randomly generated weights such that $w_{1}^{*}>w_{2}>w_{3} \in [0,1]$ and $w_{1}^{*} + w_{2} + w_{3} = 1$ as follows :
\end{itemize}
$$\forall x_{i}^{p_{L}}\;\;  x_{i}^{p_{L}} = w_{1}^{*} \times x_{i}^{p_L^{*}} \:+\: w_{2} \times x_{i}^{p_{A}^{p_{L}}} \:+\: w_{3} \times {\frac {p_{B_1}^{p_L}+p_{B_2}^{p_L}+\cdots +p_{B_n}^{p_L}}{n-2}}, \;\;p_{L} \in \textit{\textbf{P}}_{L} $$

\begin{itemize}
\item Advocate : The search direction of every advocate $p_A \in \textit{\textbf{P}}_A$ is influenced by its corresponding leader $p_L \in \textit{\textbf{P}}_L$, every associated believer $p_B \in {\textit{\textbf{p}}}_{B}^{p_{L}}$ and the associated randomly generated weights such that $w_{1}^{*}>w_{2} \in [0,1]$ and $w_{1}^{*} + w_{2} = 1$ as follows :
\end{itemize}
$$\forall x_{i}^{p_{A}}\;\;  x_{i}^{p_{A}} = w_{1}^{*} \times x_{i}^{p_L} \:+\: w_{2} \times {\frac {p_{B_1}^{p_L}+p_{B_2}^{p_L}+\cdots +p_{B_n}^{p_L}}{n-2}}, \;\;p_{A} \in \textit{\textbf{P}}_{A} $$

\begin{itemize}
\item Believers : The search direction of every believer $p_B \in \textit{\textbf{P}}_B$ is influenced by its corresponding leader $p_L \in \textit{\textbf{P}}_L$, advocate $p_A \in {\textit{\textbf{p}}}_{A}^{p_{L}}$ and the associated randomly generated weights such that $w_{1}^{*}>w_{2} \in [0,1]$ and $w_{1}^{*} + w_{2} = 1$ as follows :
\end{itemize}
$$\forall x_{i}^{p_{B}}\;\;  x_{i}^{p_{B}} = w_{1}^{*} \times x_{i}^{p_L} \:+\: w_{2}  \times x_{i}^{p_{A}^{p_{L}}}, \;\;p_{B} \in \textit{\textbf{P}}_{B} $$

\textbf{Step 3 (Updation: Global and Local Ranking) :}
After corresponding search directions are calculated, individuals are updated with new search directions, individuals within each set are locally ranked and positions are assigned accordingly, followed by global ranking based on the fitness value of leaders of corresponding groups. Group with Global Best Leader ($p_{L^*}$) is assigned as Group1.

\textbf{Step 4 (Convergence) :}
No significant improvement in the global as well as local group leaders or maximum iterations reached. Else continue to \textbf{Step 2}

\FloatBarrier

\section{Problem description and formulations}

\subsection{Benchmark Test Problems}
The LAB was tested by solving 27 well-studied benchmark problems (Table \ref{tab:Functions})\cite{KARABOGA2009108,ABC2}. The results are compared with contemporary algorithms.\\
\pagebreak

\begin{longtable}{l l c c c c}
\captionsetup{singlelinecheck = false, format= hang, justification=raggedright, font=footnotesize, labelsep=space}
\caption{The benchmark test problems (Low/Lower and Up/Upper Bounds; S = Separable; U = Unimodal; N = Non-separable;M = Multimodal).}\\
\hline
\textbf{Function} & \textbf{Name} & \textbf{Type} & \textbf{Low}  & \textbf{Up} & \textbf{Dimension} \\
\hline
\endfirsthead
\multicolumn{6}{c}%
{\tablename\ \thetable\ -- \textit{Continued from previous page}} \\
\hline
\textbf{Problem} & \textbf{Name} & \textbf{Type} & \textbf{Low}  & \textbf{Up} & \textbf{Dimension} \\
\hline
\endhead
\hline \multicolumn{6}{r}{\textit{Continued on next page}} \\
\endfoot
\hline
\endlastfoot
F1 & Foxholes & MS & {-65.536} & 65.536 & 2 \\  F5 & Ackley & MN & −32 & 32 & 30 \\ 
F7 & Bohachecsky1 & MS & −100 & 100 & 2 \\ F8 & Bohachecsky2 & MN & −100 & 100 & 2 \\ F9 & Bohachecsky3 & MN & −100 & 100 & 2\\
F10 & Booth & MS & −10 & 10 & 2 \\ 
F13 & Dixon-Price & UN & −10 & 10 & 30 \\ F15 & Fletcher & MN & −3.1416 & 3.1416 & 2 \\
F16 & Fletcher & MN & −3.1416 & 3.1416 & 5 \\ F17 & Fletcher & MN & −3.1416 & 3.1416 & 10 \\ F18 & Griewank & MN & −600 & 600 & 30 \\ 
F19 & Hartman3 & MN & 0 & 1 & 3 \\  F20 & Hartman6 & MN & 0 & 1 & 6 \\ F21 & Kowalik & MN & -5 & 5 & 4 \\
F23 & Langermann5 & MN & 0 & 10 & 5 \\ F24 & Langermann10 & MN & 0 & 10 & 10 \\ 
F25 & Matyas & UN & −10 & 10 & 2 \\ F32 & Quartic & US & −1.28 & {1.28} & 30 \\ 
F33 & Rastrigin & MS & −5.12 & 5.12 & 30 \\ F35 & Schaffer & MN & −100 & 100 & 2 \\
F37 & Schwefel\_1\_2 & UN & −100 & 100 & 30 \\ F38 & Schwefel\_2\_22 & UN & −10 & 10 & 30 \\
F43 & Six-hump camelback & MN & −5 & 5 & 2 \\ F44 & Sphere2 & US & −100 & 100 & 30 \\
F45 & Step2 & US & −100 & 100 & 30 \\ F47 & Sumsquares & US & −10 & 10 & 30 \\
F50 & Zakharov & UN & −5 & 10 & 10
\label{tab:Functions}
\end{longtable}

\subsection{Manufacturing And Machining Problems}
Engineering problems are generally complex in nature and may involve several local optima. The complexity grows when the associated objective function involves coupled variables. This necessitates development of approximation algorithms, which can efficiently jump out of local optima and search for the global optimum \cite{GANDOMI2020112917,LEE20053902}. The LAB algorithm’s performance was tested by solving three types of engineering problems in the domain of machining, namely Abrasive Water Jet Machining, Electric Discharge Machining, micro-machining and Process Parameter Optimization for Turning of Alloy.

\subsubsection{Abrasive Water Jet Machining (AWJM)}

AWJM is an an extended version of water jet cutting, which uses water as the material to impinge on the work material to result in a cut. It can be also used for machining a heat-sensitive materials, as the heat generated is very low as well as the cut is 10x times faster than conventional methods.

Four critical parameters are $u_2$ (in $mm$) as nozzle diameter, $u_3$ (in $mm$) as standoff distance, $u_4$ (in $mm/min$) as cutting head speed/traverse speed and $u_1$ (in $mm$) as workpiece thickness \cite{Kecha,SHANMUGAM2008923,SHUKLA2017212}
for which the associated responses are surface roughness $R_a$ and taper angle $kerf$\cite{dhanawade}. It is evident all the process paramters interact with one another, affecting precision of cuts. Hence, optimum combination of the above process parameters is required for optimum results. Formulated regression model of the AWJM process is adopted here\cite{Kecha,SHUKLA2017212}. The function being linear nonseperable makes it complex to solve and increases the chances of getting stuck in the local minima making the problem harder and tedious to solve. The formulated regression model adopted is as shown below:

\begin{equation}
\begin{split}\label{Ra_AWJM}
\textrm{Minimize} \;\;  R_a = -23.309555 + 16.6968 u_{1} + 26.9296  u_{2} + 0.0587 u_{3} + 0.0146  u_{4} - 5.1863  u_{2}^2\\ 
- 10.4571 u_{1}  u_{2} - 0.0534 u_{1} u_{3} - 0.0103 u_{1} u_{4} + 0.0113  u_{2} u_{3} - 0.0039  u_{2} u_{4}
\end{split} 
\end{equation}

\begin{equation}
\begin{split}\label{kerf_AWJM}
\textrm{Minimize} \;\; kerf = -1.15146+0.70118 u_1+2.72749 u_2+0.00689 u_3-0.00025 u_4\\
+0.00386 u_2 u_3-0.93947 u_2^2-0.25711 u_1 u_2-0.00314 u_1 u_3\\
-0.00249 u_1 u_4+0.00196 u_2 u_4-0.00002 u_3 u_4-0.00001 u_3^2
\end{split} 
\end{equation}

where\;
$0.9 \leq u_1 \leq 1.25$,\;
$0.95 \leq u_2 \leq 1.5$,\;
$20 \leq u_3 \leq 96$,\;
$200 \leq u_4 \leq 600$

\subsubsection{Electric Discharge Machining (EDM)}

One of the elctro-thermal non-traditional machining processes is the EDM, which uses electrical spark or thermal energy to erode unwanted material in order to create desired shape. It is a controlled metal-removal process that is used to remove metal by means of electric spark erosion. The metal-removal process is performed by applying a pulsating (ON/OFF) electrical charge of high-frequency current through the electrode to the workpiece. In the gap between the tool and the workpiece, a difference in the applied potential is formed, establishing an electric field. Due to which the loose electrons on the tool gain high velocity and energy when subjected to electrostatic forces, after which these free electrons collide with the dielectric molecules which results in ionization. More the electrons get accelerated, more positive ions and electrons get generated resulting in increase in the concentration of electrons and ions. The energy released causes electrode wear rate to take place\cite{muthu,gopal} resulting in case hardening of the workpiece.\\

In order to control surface roughness $R_a$ process parameters: $v_1$ (in $A$) as discharge current, $v_2$ (in $V$) as gap voltage, $v_3$ (in $\mu s$) as pulse on-time and $v_4$ (in $\mu s$) as pulse off-time need to be optimized for the EDM process. The process responses for surface finish and electrode wear rate of machined component are $MRR$, $R_a$ and relative electrode wear rate $REWR$, respectively.
The regression model for the above process is adopted here\cite{SHUKLA2017212,TzengCJ}:
\begin{equation}
\begin{split}\label{MRR_EDM}
\textrm{Maximize} \;\; MRR = -235.15 + 39.7v_1 + 4.277v_2 + 1.569v_3 - 1.375v_4 - 0.0059v_3^2 - 0.536v_1v_2
\end{split} 
\end{equation}
\begin{equation}
\begin{split}\label{Ra_EDM}
\textrm{Minimize} \;\; R_a = 30.347 - 0.618v_1 - 0.438v_2 + 0.059v_3 - 0.59v_4 + 0.019v_1v_4 + 0.0075v_2v_4
\end{split} 
\end{equation}
\begin{equation}
\begin{split}\label{REWR_EDM}
\textrm{Minimize} \;\; REWR = 196.564 - 24.19v_1 - 3.135v_2 - 1.781v_3 + 0.153v_4 + 0.464v_1v_2 + 0.158v_1v_3\\ 
+ 0.025v_1v_4 + 0.029v_2v_3 - 0.017v_2v_4 - 0.003385v_1v_2v_3+0.093v_1^2 \\
+ 0.001491v_3^2 + 0.005265v_4^2
\end{split} 
\end{equation}

where $\;
7.5 \leq v_1 \leq 12.5, \;
45 \leq v_2 \leq 55, \;
50 \leq v_3 \leq 150, \;
40 \leq v_4 \leq 60$

\subsubsection{Micro-machining processes}
The various processes of cutting raw materials into specific dimensions in a controlled removal process is termed as machining. 
This process of machining usually consists of a cutting tool, machine tool and a workpiece\cite{EZUGWU20051353}. Machinability refers to evaluation of ease for cutting any type of material in minimum cost and time into a specific shape and dimension for a certain tolerance, surface quality, etc.,\cite{DeGarmo}.

Micro-turning is a type of micro-machining process which uses solid micro-tools to remove material from workpiece and is almost similar to conventional turning operation. 
The micro-tools used to remove workpiece material have significant characteristics which significantly affect the size reduction\cite{Qin2010}.

$w_1$ (in $m/min$) as cutting speed , $w_2$ (in $\mu/rev$) as feed and $w_3$ (in $\mu m$) as depth of cut are the process parameters for micro-turning. Performance responses are flank wear ($f_b$) and surface roughness ($R_a$).
The formulated regression model for the above process is adopted here\cite{DURAIRAJ2013878} :

\begin{equation}
\begin{split}\label{fb_6}
\textrm{Minimize} \;\; f_b = 0.004 w_1^{0.495} w_2^{0.545} w_3^{0.763}
\end{split} 
\end{equation}
\begin{equation}
\begin{split}\label{Ra_7}
\textrm{Minimize} \;\; R_a = 0.048w_1^{-0.062}w_2^{0.445}w_3^{0.516}
\end{split} 
\end{equation}

where $25 \leq w_1 \leq 37,\; 5 \leq w_2 \leq 15,\; 30 \leq w_3 \leq 70 $

Process parameters $x_1$ (in $rpm$) as cutting speed, $x_2$ (in $mm/min$) as feed and process responses surface roughness $R_a$ and machining time $M_t$ with two milling cutters with diameters $0.7 mm$ and $1 mm$ are considered for micro-milling process \cite{aniket}. The formulated regression model for the above process is adopted here:

Tool with diameter $0.7 mm$
\begin{equation}
\begin{split}\label{Ra_8}
\textrm{Minimize} \;\; R_a = -0.455378 + 0.00027f_1 + 0.16422f_2 - 0.000077f_1f_2 
\end{split} 
\end{equation}
\begin{equation}
\begin{split}\label{Mt_9}
\textrm{Minimize} \;\; M_t = 17.71644 - 0.0002f_1 - 4.8404f_2 + 0.0001f_1f_2
\end{split} 
\end{equation}

Tool with diameter $1 mm$ 
\begin{equation}
\begin{split}\label{Ra_10}
\textrm{Minimize} \;\; R_a = -0.208871 + 0.000144f_1 + 0.019571f_2 
\end{split} 
\end{equation}
\begin{equation}
\begin{split}\label{Mt_11}
\textrm{Minimize} \;\; M_t = 20.2906 - 0.0015f_1 - 5.8369f_2 + 0.0006f_1f_2 
\end{split} 
\end{equation}

where $\; 1500 \leq f_1$ $\leq 2500,\; 1 $ $\leq f_2 \leq 3$\\

$B_h$ as Burr height and $B_t$ as burr thickness, are performance responses for four drilling cutter diameters in micro-drilling process, $0.5 mm$; $0.6 mm$; $0.8 mm$ and $0.9 mm$. Process parameters for the above are $y_1$ (in $rpm$) as cutting speed and $y_2$ (in $mm/min$) as feed. The formulated regression model of the above process is adopted here for the tools used as follows\cite{pansari}

Tool with diameter $0.5 mm$
\begin{equation}
\begin{split}\label{Bh_12}
\textrm{Minimize} \;\; B_h = 420.94 - 0.234g_1 - 99.91g_2 + 6.55 \times 10^-5g_1^2 + 22.152g_2^2 
\end{split} 
\end{equation}
\begin{equation}
\begin{split}\label{Bt_13}
\textrm{Minimize} \;\; B_t = 90.57 - 0.049g_1 - 27.12g_2 + 1.32 \times 10^{-5}g_1^2 + 5.54g_2^2  
\end{split} 
\end{equation}

Tool with diameter $0.6 mm$
\begin{equation}
\begin{split}\label{Bh_14}
\textrm{Minimize} \;\; B_h = 369.67 - 0.028g_1 - 156.79g_2 + 6.64 \times 10^-6g_1^2 + 23.162g_2^2 
\end{split} 
\end{equation}
\begin{equation}
\begin{split}\label{Bt_15}
\textrm{Minimize} \;\; B_t = 35.34 - 0.019g_1 - 0.59g_2 + 6.44 \times 10^{-6}g_1^2 + 0.51g_2^2 
\end{split} 
\end{equation}

Tool with diameter $0.8 mm$
\begin{equation}
\begin{split}\label{Bh_16}
\textrm{Minimize} \;\; B_h = 106.116 + 0.13g_1 - 6.62g_2 + 1.49 \times 10^{-6}g_1^2 + 4.75g_2^2 
\end{split} 
\end{equation}
\begin{equation}
\begin{split}\label{Bt_17}
\textrm{Minimize} \;\; B_t = 59.79 - 0.024g_1 - 11.3g_2 + 7.78 \times 10^{-6}g_1^2 + 2.18g_2^2 
\end{split} 
\end{equation}

Tool with diameter $0.9 mm$
\begin{equation}
\begin{split}\label{Bh_18}
\textrm{Minimize} \;\; B_h = 450.7 - 0.09g_1 - 34.48g_2 + 2.34 \times 10^{-5}g_1^2 + 5.03g_2^2 
\end{split} 
\end{equation}
\begin{equation}
\begin{split}\label{Bt_19}
\textrm{Minimize} \;\; B_t = 80.07 - 0.040g_1 - 14.81g_2 + 1.516 \times 10^{-5}g_1^2 + 4.65g_2^2 
\end{split} 
\end{equation}

where$\; 1000 \leq g_1 \leq 2500,\; 1 \leq g_2 \leq 4$

\subsubsection{Process parameter optimization for turning of titanium alloy (MQL environment)}

Minimum Quantity Lubrication (MQL) has increasingly been adopted over the past few years in the manufacturing domain, due to its abilities to reduce costs and material wastes as compared with traditional methods. In MQL, a small quantity of cutting fluid such as sustainable lubricants (vegetable oil) is applied on the tool-chip surface region as well as compressed air acting as an alternative for coolant fluids. Thus cutting costs by avoiding use of huge amounts of coolant fluids, thus focusing more on the heat generated rather than using coolants to reduce surface temperatures resulting in increase in tool life\cite{GUPTA201767}. 

$k_1 (mm/rev)$ as Feed rate, $k_2$(degrees) as approach angle, $k_3 (m/min)$ as cutting speed are considered process parameters and the performance responses for the above are $F_c$ as tangential force, $V_{Bmax}$ as tool wear, $R_a$ as surface roughness and $L$ as tool-chip contact length. The formulated regression model of the above process is adopted here\cite{Gupta2016,SNNiknam}:

\begin{equation}
\begin{split}\label{MQL_Fc}
\textrm{Minimize} \;\; F_c = -202.01471 + 1.28250 \times k_3 + 3225 \times k_1 - 0.74167 \times  k_2 - 9.4 \times k_3 \times k_1 
\end{split} 
\end{equation}
\begin{equation}
\begin{split}\label{MQL_VBmax}
\textrm{Minimize} \;\; V_{Bmax} = -0.27368 + 0.001575 \times k_3 + 2.4 \times k_1 - 0.0010833 \times k_2 
\end{split} 
\end{equation}
\begin{equation}
\begin{split}\label{MQL_Ra}
\textrm{Minimize} \;\; R_a = -0.16294 + 0.001425 \times k_3 + 3.7 \times k_1 - 0.000416667 \times k_2
\end{split} 
\end{equation}
\begin{equation}
\begin{split}\label{MQL_L}
\textrm{Minimize} \;\; L = 0.96302 - 0.00215931 \times k_3 + 0.92703 \times k_1 + 0.00152807 \times k_2 
\end{split} 
\end{equation}

where $\;200 \leq k_1 \leq 300, \;0.1 \leq k_2 \leq 0.2, \;60 \leq k_3 \leq 90$

\section{Tests and Validations}
The LAB algorithm was coded in Python3 on Google Collab Platform with an Intel(R) Xeon(R) @2.30 GHz Intel Core 2 Duo processor with 12 GB RAM. In the initialization step, individuals were generated and randomly assigned to groups. The selected LAB parameters are: number of groups $G=4$, number of individuals in each group $n=5$, max iterations = 100.

\subsection{Benchmark Problems}

LAB is validated by solving 27 benchmark test functions and the results are compared with other algorithms which are necessarily stochastic in nature.
The criteria for comparison are mean and best solutions, standard deviation as well as runtime of the algorithms(Refer to Table \ref{tab:TestTable}).

A statistical analysis is performed by executing two-sided and pairwise Wilcoxon signed-rank test (Refer to Tables \ref{tab:StatTable} and \ref{tab:multiprob}). In the two-sided comparison optmimum solutions obtained from 30 independent runs solving a benchmark test problem using LAB are compared with other algorithms solving the same benchmark test problems.
Significance value $\alpha$ was chosen as 0.05 with a null hypothesis H0: the median of solutions obtained by algorithms A and B are equal, in order to verify if an alternative hypothesis exists i.e. performance of algorithm B is better than algorithm A or the other way around, the size of the ranks provided by Wilcoxon signed-rank (T+ and T- values) were thoroughly examined \cite{CIVICIOGLU20138121}.

At the bottom of the Table \ref{tab:StatTable} counts of significant cases (+/-/=) are mentioned. The results obtained exhibited the superior performance of LAB algorithm as compared to the other algorithms. In the pairwise comparison, the average of the best solutions obtained by the algorithms over 30 runs for solving the benchmark test problems is compared.

\begin{figure}[!htb]
    \centering
    \includegraphics[width=0.99\linewidth]{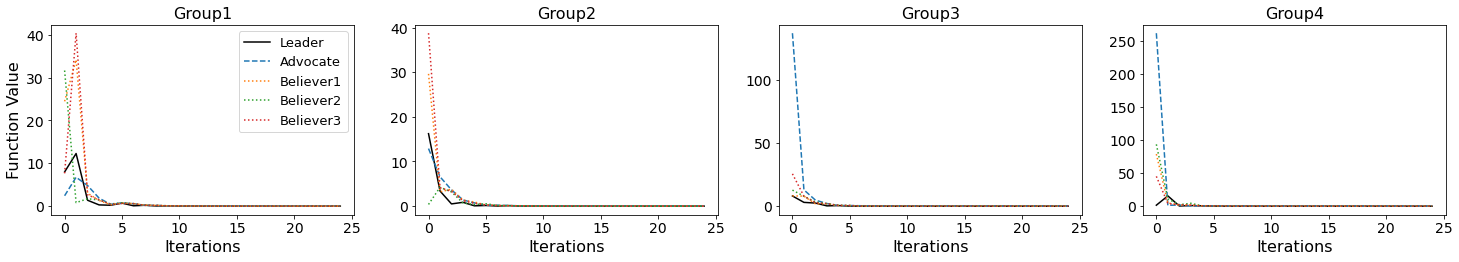}
    \caption{Convergence: Booth Function(F10)}
    \label{fig:booth}
\end{figure}

\begin{figure}[!htb]
    \centering
    \includegraphics[width=0.99\linewidth]{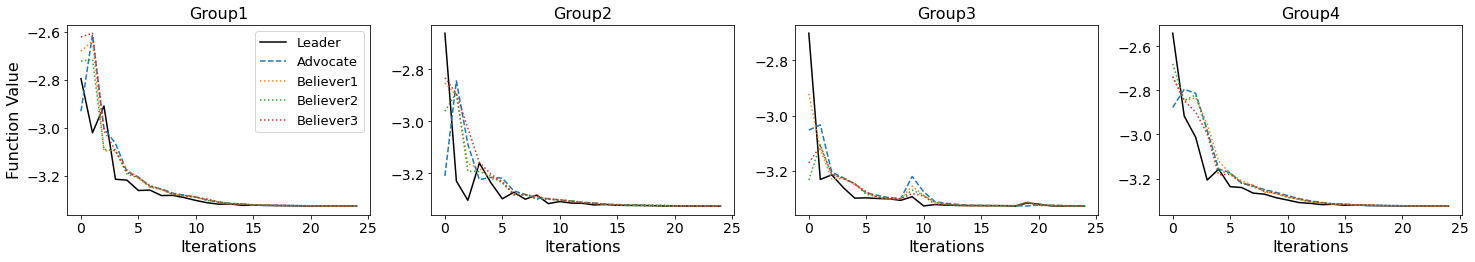}
    \caption{Convergence: Hartmann6 Function(F20)}
    \label{fig:hartmann6}
\end{figure}

\begin{figure}[!htb]
    \centering
    \includegraphics[width=0.99\linewidth]{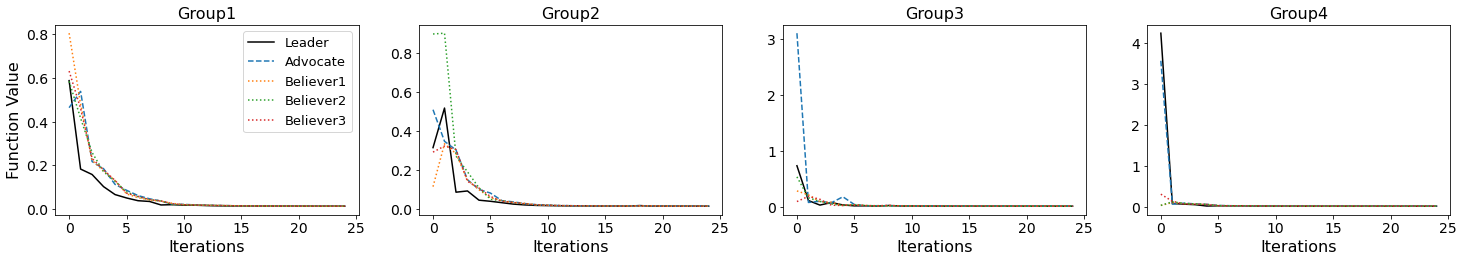}
    \caption{Convergence: Matyas Function(F25)}
    \label{fig:matyas}
\end{figure}

\begin{figure}[!htb]
    \centering
    \includegraphics[width=0.99\linewidth]{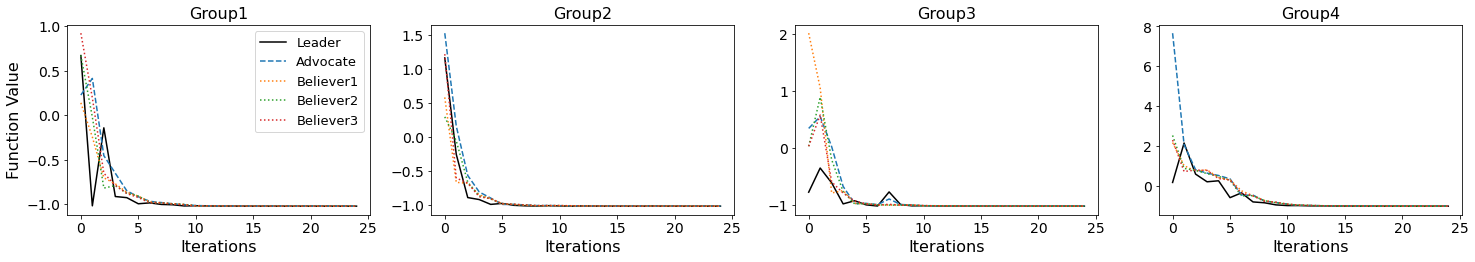}
    \caption{Convergence: Six-hump Camelback Function(F43)}
    \label{fig:sixhump}
\end{figure}

The convergence plots of few selected functions namely Booth(unimodal), Hartmann6(unimodal), Matyas(multimodal) and Six-hump camelback(multimodal) are presented in Figures \ref{fig:booth}–\ref{fig:sixhump}.
These plots exhibit the competitive behaviour of individuals within a group to reach optimum solution. It is also evident from the plots that the individuals in the group follow the leader. During every iteration, group leaders from corresponding groups compete with the global best leader and are successful at times and the group associated with the global best leader is assigned as Group 1. Thus, changing the search direction of the local leaders following the global best leader. The abrupt changes in the graph exhibit this phenomena of competitiveness of individuals. The convergence highlights the significance of LAB approach by quickly reaching the optimum solution.

\begin{landscape}
    \begin{table}[!htb]
    \captionsetup{singlelinecheck = false, format= hang, justification=raggedright, font=footnotesize, labelsep=space}
    \caption{Statistical solutions of algorithms for Benchmark test problems\\(\textit{Mean= Mean solution;Std. Dev.= Standard Deviation;Best= Best Solution; Runtime= Mean Runtime in Seconds})}
    \includegraphics[width=0.85\linewidth]{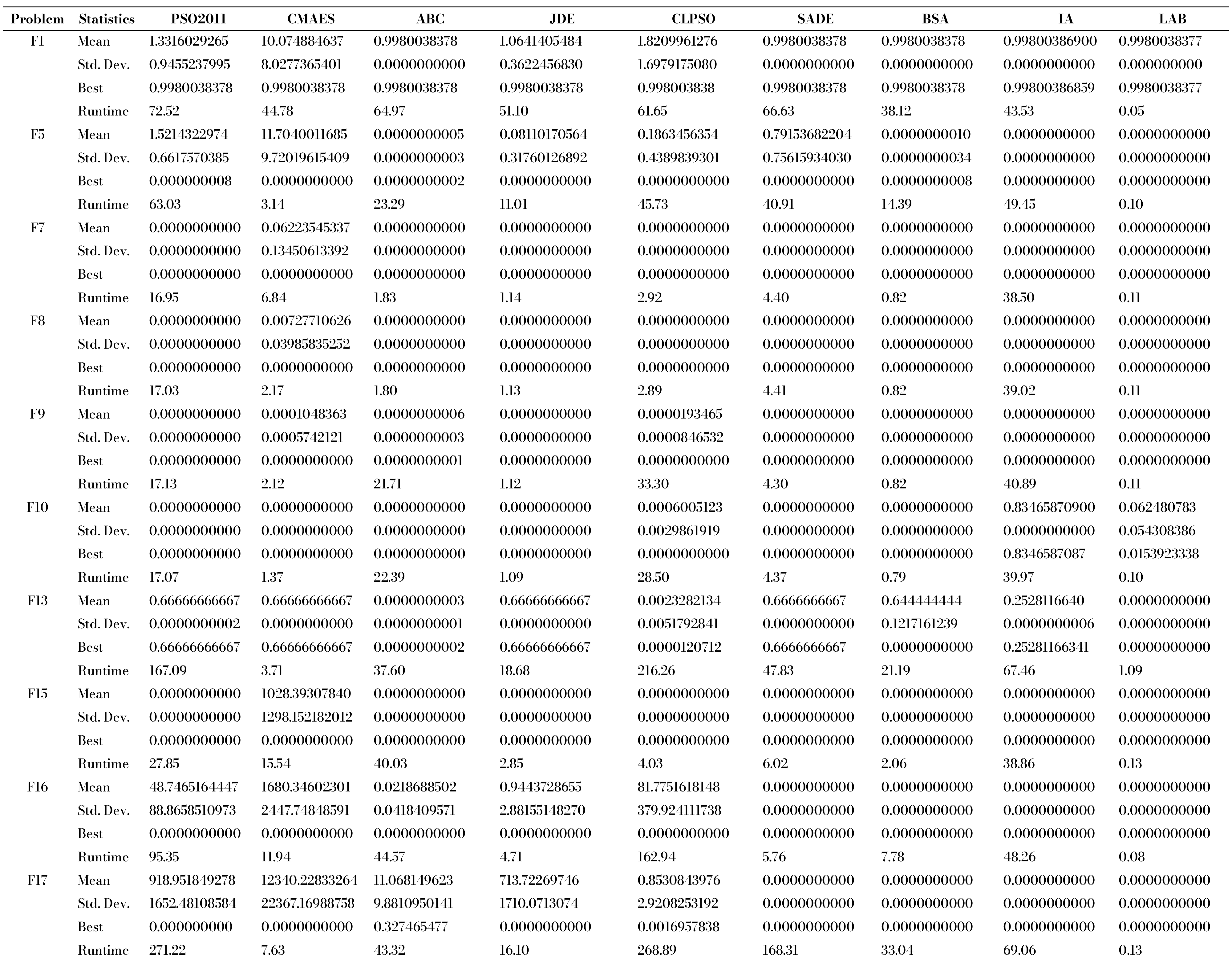}
    \centering
    \label{tab:TestTable}
    \end{table}
    
    \begin{table}[!htb]
    \captionsetup{singlelinecheck = false, format= hang, justification=raggedright, font=footnotesize, labelsep=space}
    \caption*{\textbf{Table \ref{tab:TestTable} Continued}}
    \includegraphics[width=0.85\linewidth]{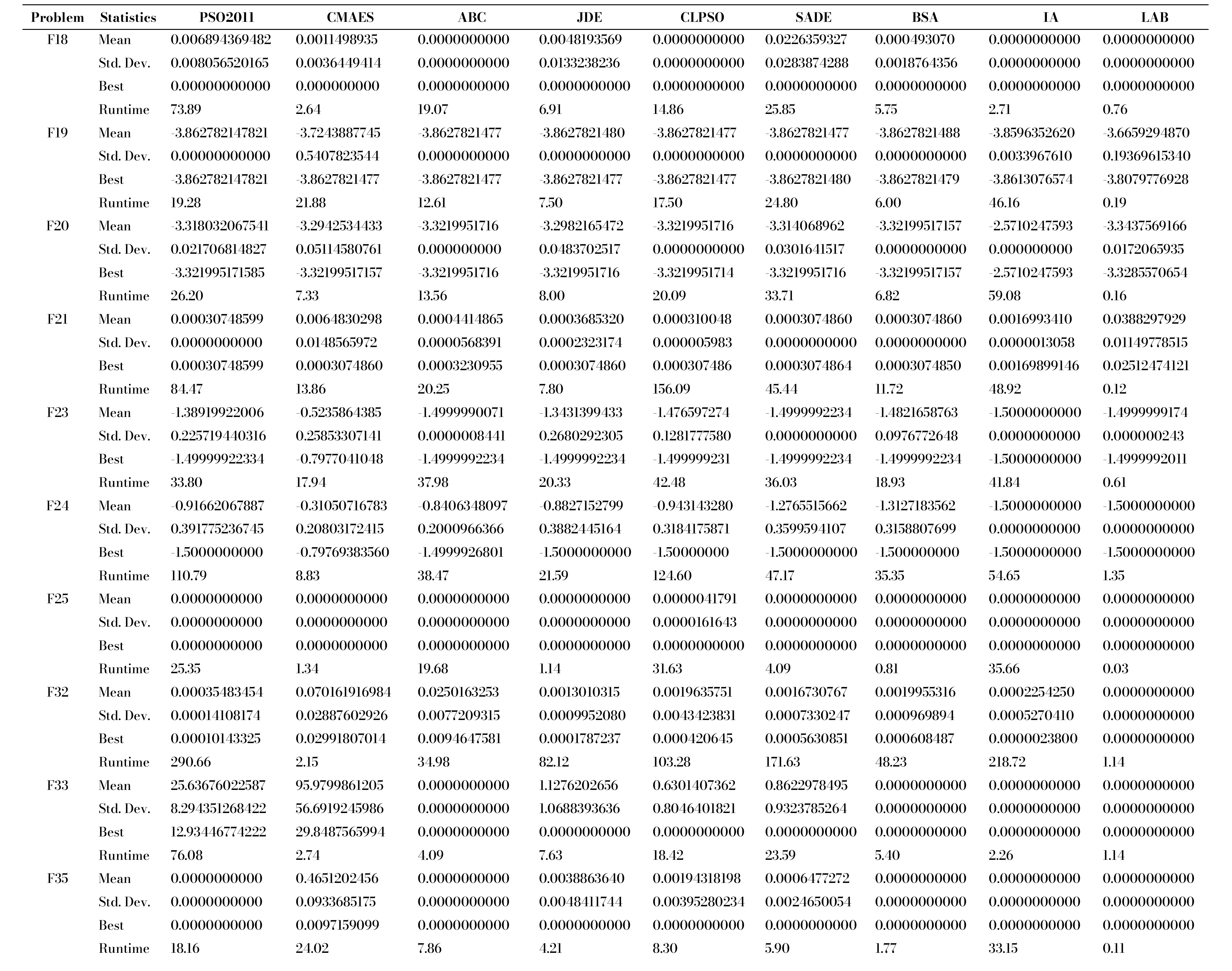}
    \centering
    \end{table}
    
    \begin{table}[!htb]
    \captionsetup{singlelinecheck = false, format= hang, justification=raggedright, font=footnotesize, labelsep=space}
    \caption*{\textbf{Table \ref{tab:TestTable} Continued}}
    \includegraphics[width=0.85\linewidth]{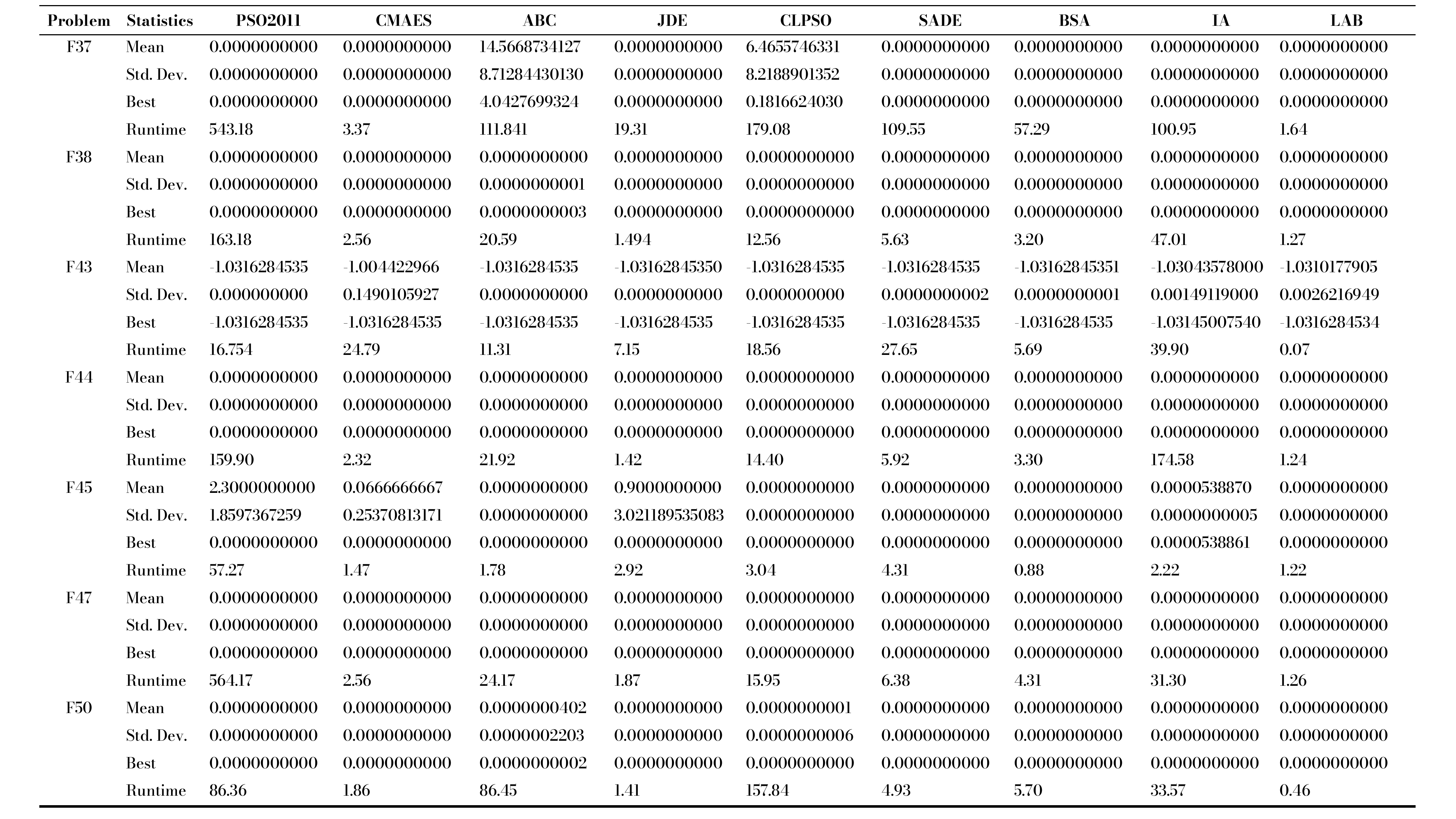}
    \centering
    \end{table}
    
\end{landscape}

\begin{table}[H]
\captionsetup{singlelinecheck = false, format= hang, justification=raggedright, font=footnotesize, labelsep=space}
\caption{Statistical results for Benchmak Test problems using two-sided Wilcoxon signed-rank test ($\alpha$ = 0.05)}
\includegraphics[width=0.99\linewidth]{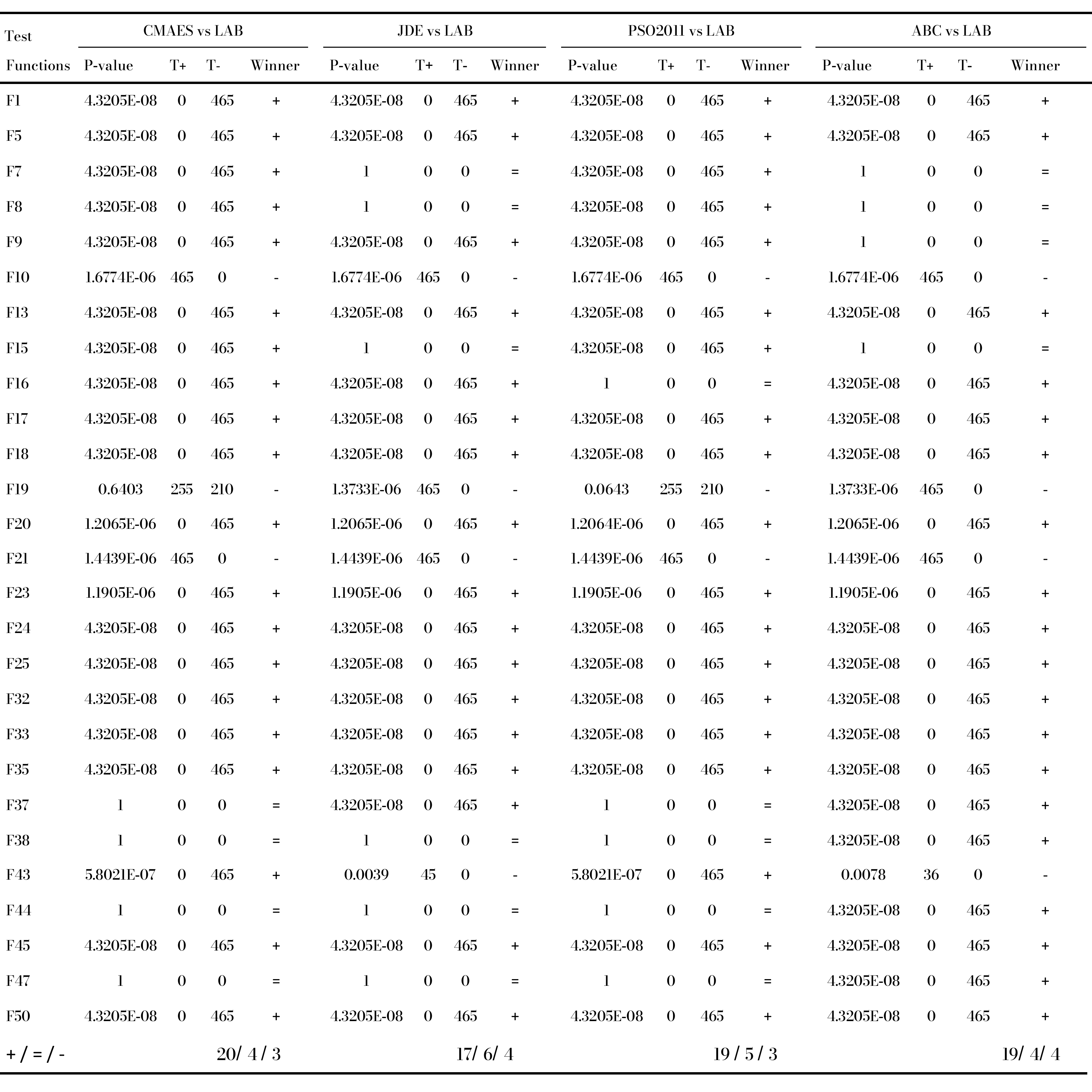}
\centering
\label{tab:StatTable}
\end{table}

\begin{table}[H]
\captionsetup{singlelinecheck = false, format= hang, justification=raggedright, font=footnotesize, labelsep=space}
\caption*{\textbf{Table \ref{tab:StatTable} Continued}}
\includegraphics[width=0.98\linewidth]{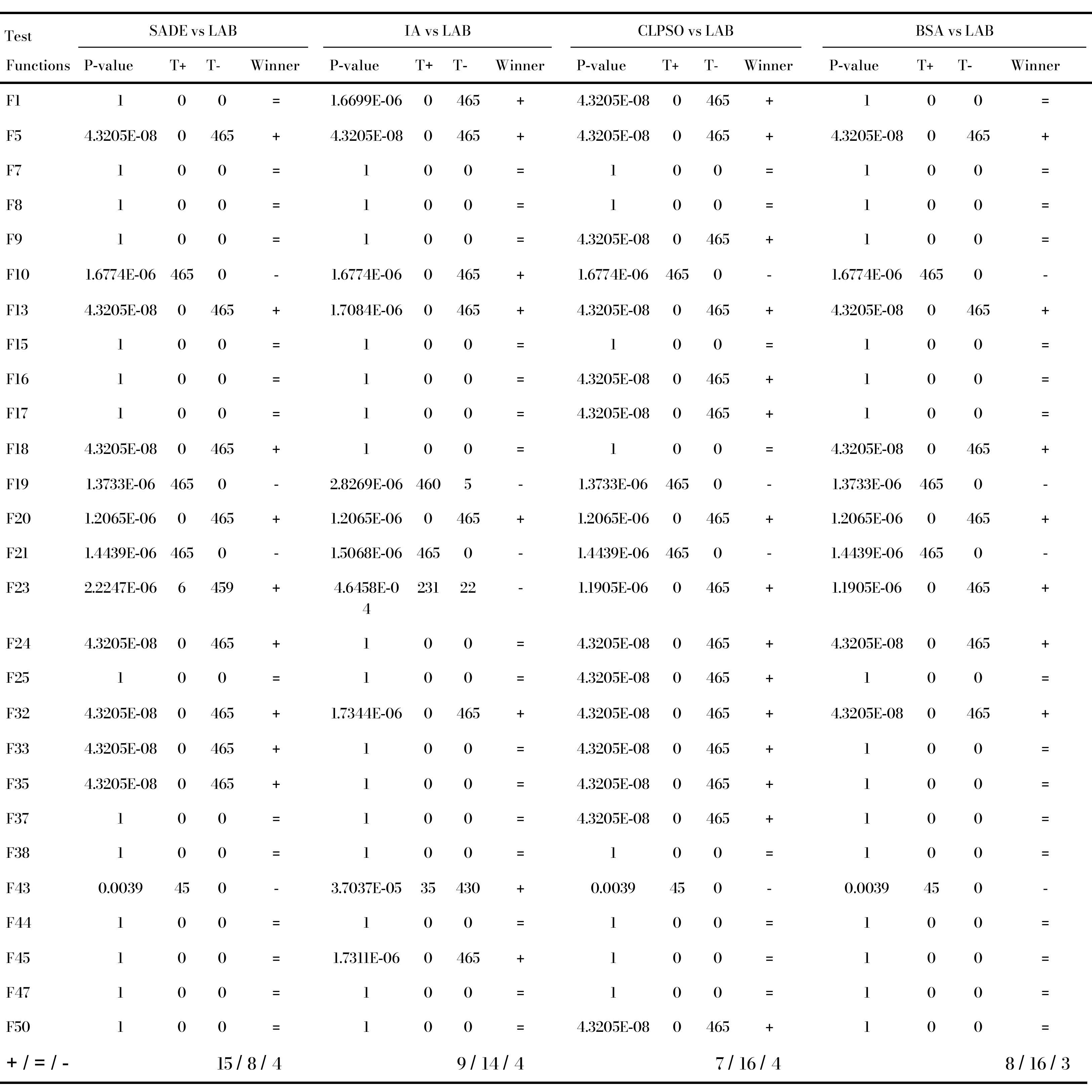}
\centering
\end{table}

\begin{table}[H]
\captionsetup{singlelinecheck = false, format= hang, justification=raggedright, font=footnotesize, labelsep=space}
\caption{Statistical pairwise comparison.}
\includegraphics[width=0.80\linewidth]{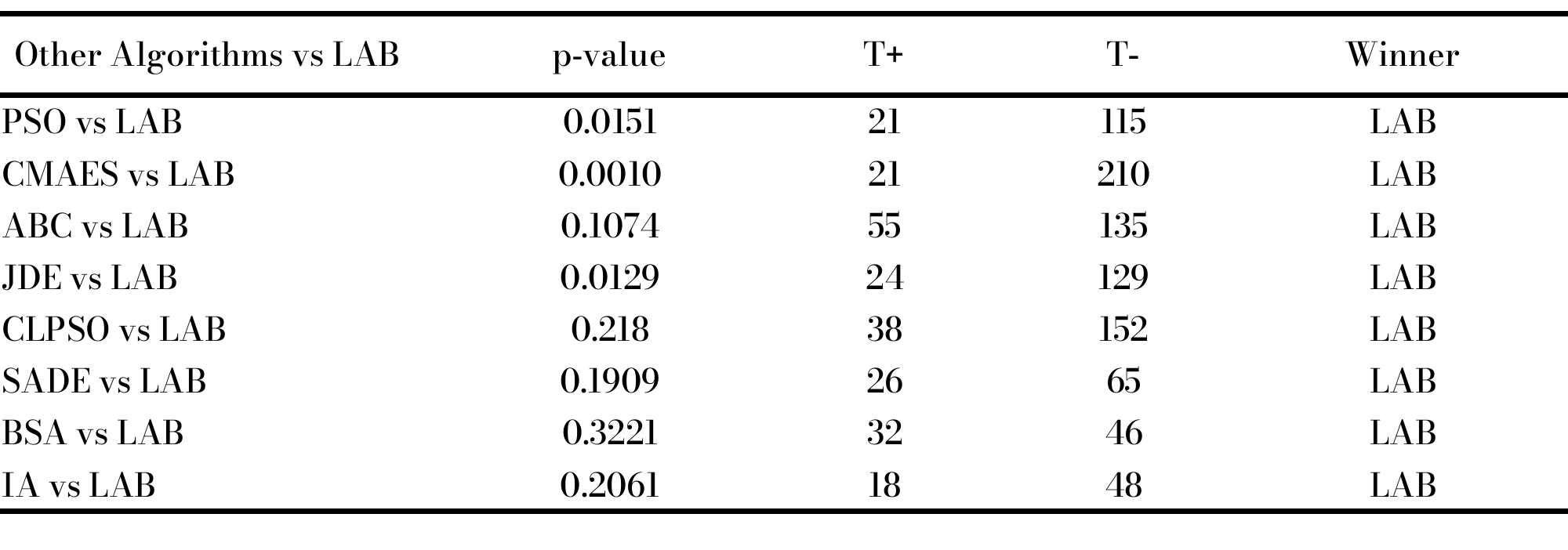}
\centering
\label{tab:multiprob}
\end{table}

\FloatBarrier

\subsection{Solutions to AWJM and EDM}

Table \ref{tab:AWJM1} contains best and mean solutions along with their associated standard deviation obtained for $R_a$ and $kerf$ of AWJM using LAB, Multi-CI, GA, SA and PSO and comparison with the variations of CI is shown in Table \ref{tab:AWJM2}. In LAB approach, individuals are randomly assigned the group and the associated leader in the first iteration. For every following iteration, the individuals follow the local best individual/leader and the local best individual/local leader also follows the global best individual/Global Leader, helps to explore better solutions due to which individuals avoided local minima. Hence, LAB yielded in better solutions as compared with FA, experimental, regression approach and PSO for $kerf$. \\
LAB was able to outperform RSA, BPNN, FA, $f_{best}$, $f_{better}$ and alienation in the matter of quality of solution for solving $MRR$ for EDM problems due to its strong exploration and exploitation mechanism evident from Table \ref{tab:CompareEDM1}.\\

It is evident in Table \ref{tab:AWJM1}, the results shown for LAB are less robust as compared to GA and Multi-CI. As compared to SA and PSO, LAB outperformed by achieving 8\% and 23\% minimization of $kerf$ in AWJM as is evident in Tables \ref{tab:AWJM1} and \ref{tab:AWJM2}. Compared to $f_{best}$, $f_{better}$ and alienation, LAB achieved 78\%, 79\% and 47\% maximization of $MRR$ for EDM as is evident in Table \ref{tab:CompareEDM1}.

\begin{table}[!htb]
\captionsetup{singlelinecheck = false, format= hang, justification=raggedright, font=footnotesize, labelsep=space}
\caption{Solutions to $R_a$ and $kerf$ of AWJM}
\includegraphics[width=0.6\linewidth]{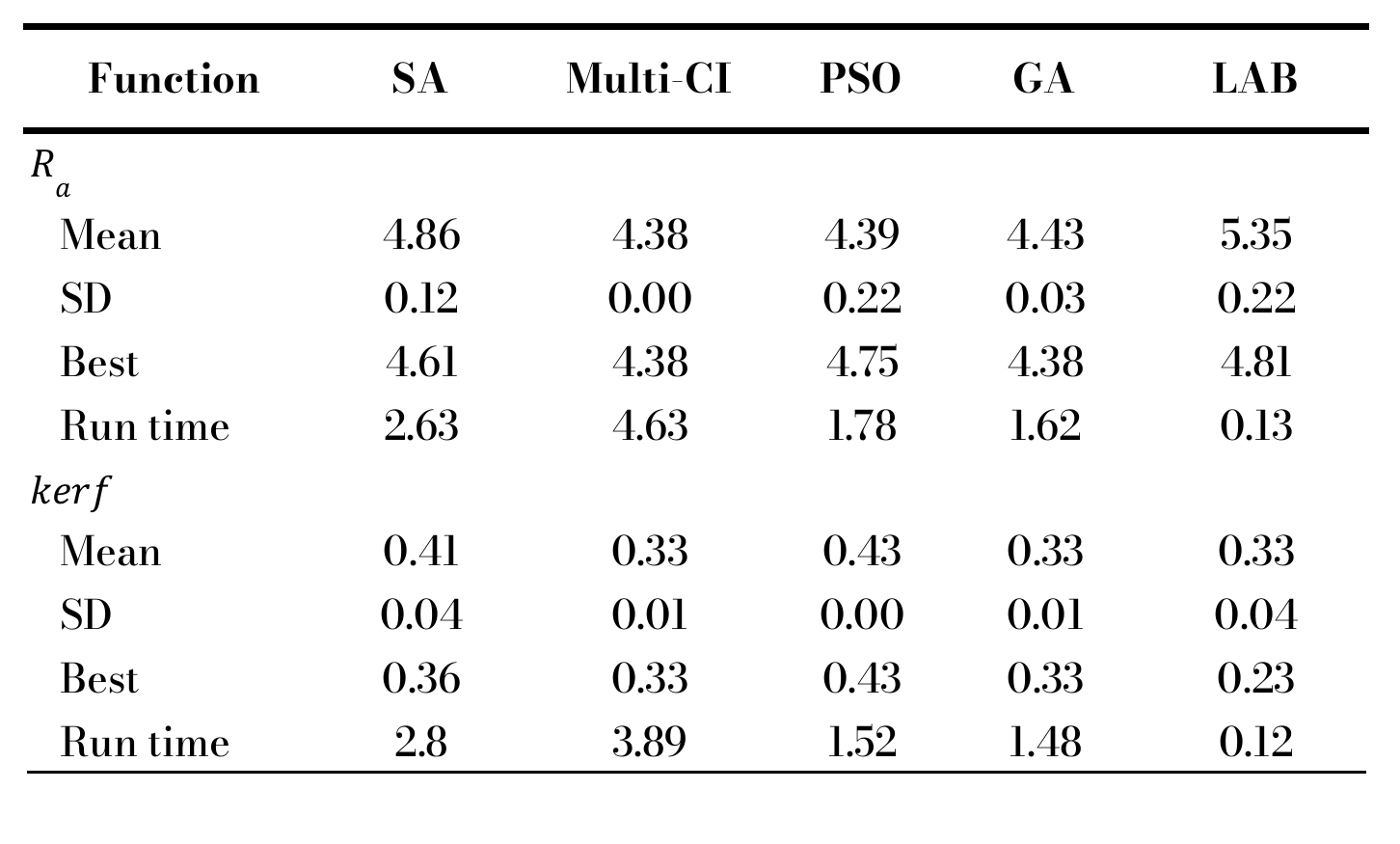}
\centering
\label{tab:AWJM1}
\end{table}

\begin{table}[!htb]
\captionsetup{singlelinecheck = false, format= hang, justification=raggedright, font=footnotesize, labelsep=space}
\caption{Overall solutions to $R_a$ and $kerf$ of AWJM}
\includegraphics[width=0.95\linewidth]{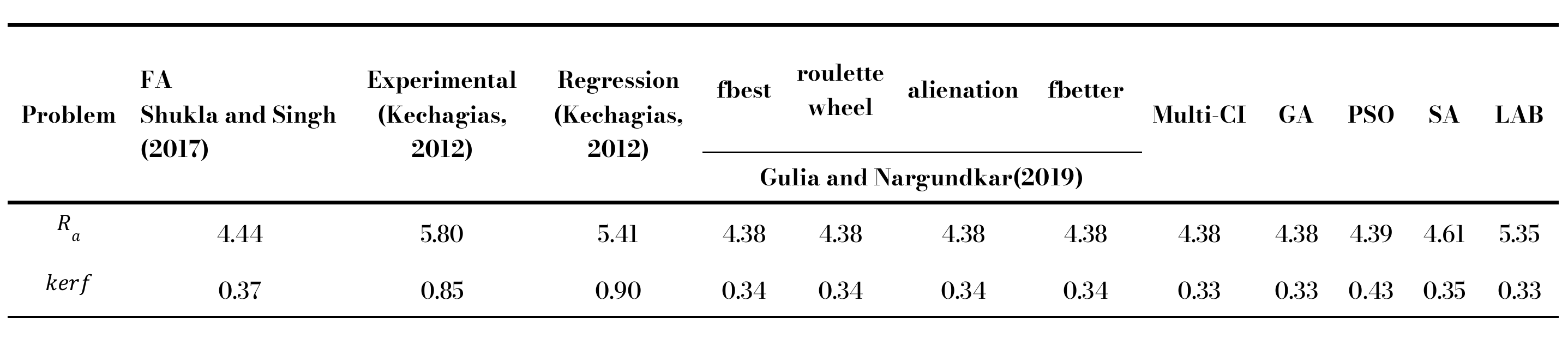}
\centering
\label{tab:AWJM2}
\end{table}

\begin{figure}
\centering
\begin{subfigure}{0.99\textwidth}
    \includegraphics[width=\textwidth]{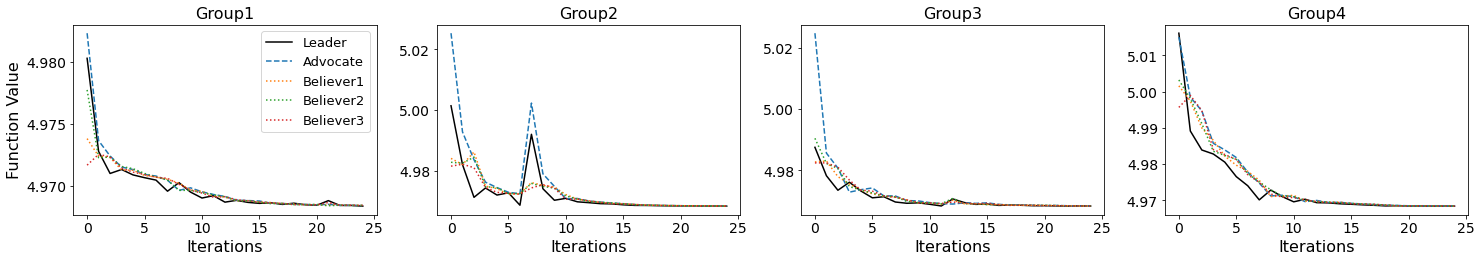}
    \caption{$R_a$}
    \label{fig:AWJM_Ra}
\end{subfigure}
\begin{subfigure}{0.99\textwidth}
    \includegraphics[width=\textwidth]{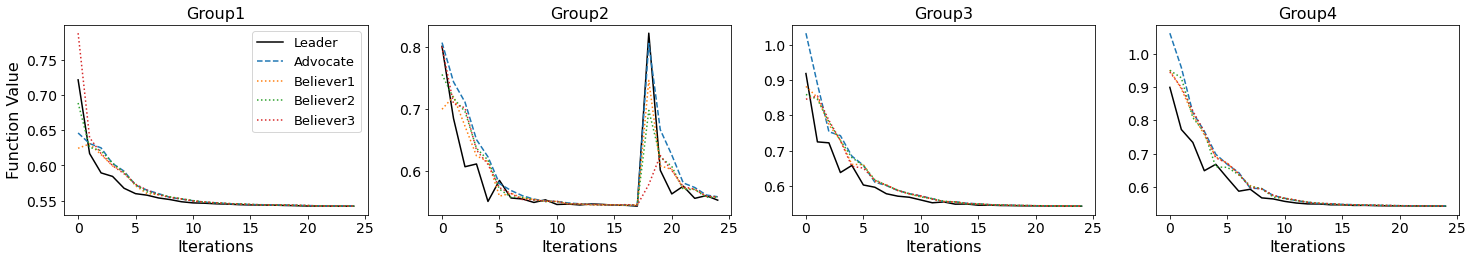}
    \caption{$kerf$}
    \label{fig:AWJM_kerf}
\end{subfigure}
\caption{Convergence: AWJM}
\label{fig:plot_AWJM}
\end{figure}

\begin{figure}[!htb]
\centering
\begin{subfigure}{0.99\textwidth}
    \includegraphics[width=\textwidth]{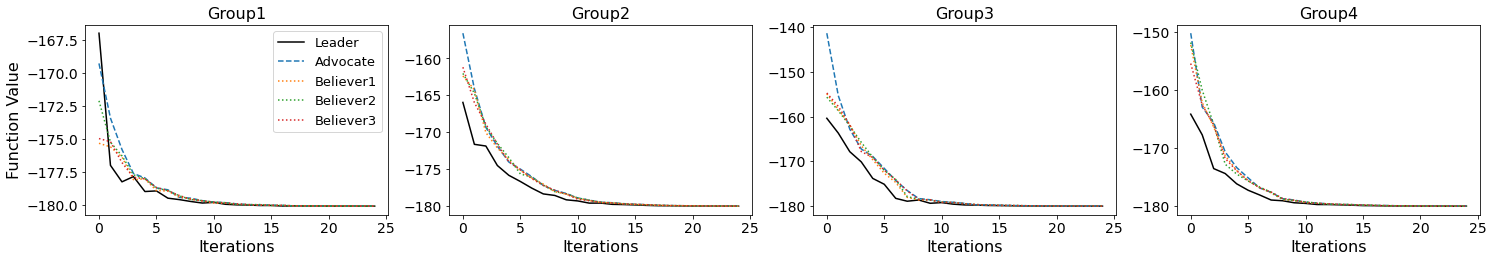}
    \caption{$MRR$}
    \label{fig:MRR}
\end{subfigure}
\begin{subfigure}{0.99\textwidth}
    \includegraphics[width=\textwidth]{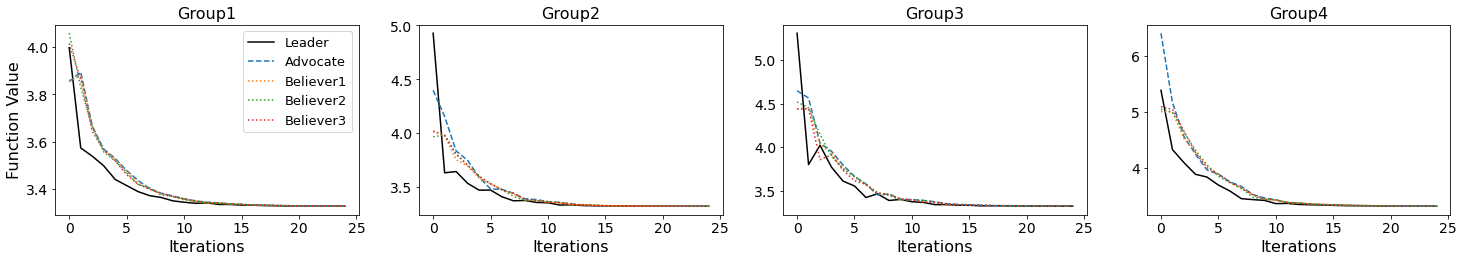}
    \caption{$R_a$}
    \label{fig:Ra_EDM}
\end{subfigure}
\begin{subfigure}{0.99\textwidth}
    \includegraphics[width=\textwidth]{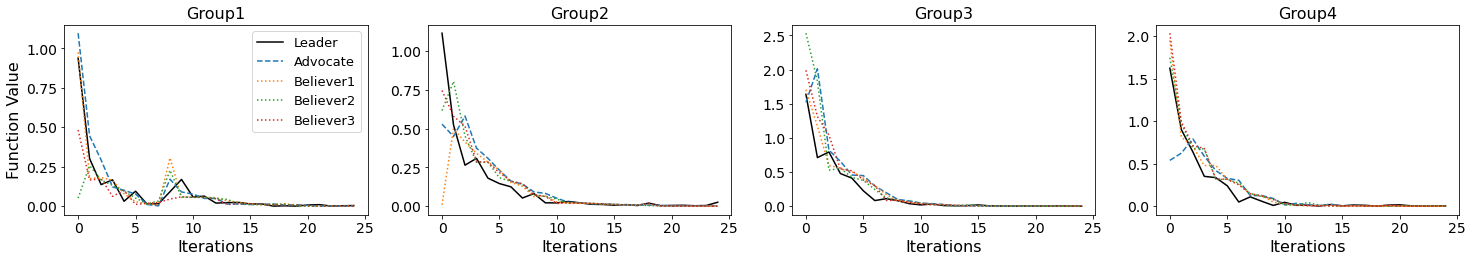}
    \caption{$REWR$}
    \label{fig:REWR}
\end{subfigure}
\caption{Convergence: EDM}
\label{fig:plot_EDM}
\end{figure}

Best solution plots in every iteration of LAB for solving AWJM and EDM problems are exhibited in Fig. \ref{fig:plot_AWJM} and Fig. \ref{fig:plot_EDM} a–c respectively, as well as the solution comparison is exhibited in Table \ref{tab:AWJM2}.

\begin{landscape}
\begin{table}[!htb]
\captionsetup{singlelinecheck = false, format= hang, justification=raggedright, font=footnotesize, labelsep=space}
\caption{Solutions to $R_a$, $MRR$, $REWR$ of EDM}
\includegraphics[width=0.96\linewidth]{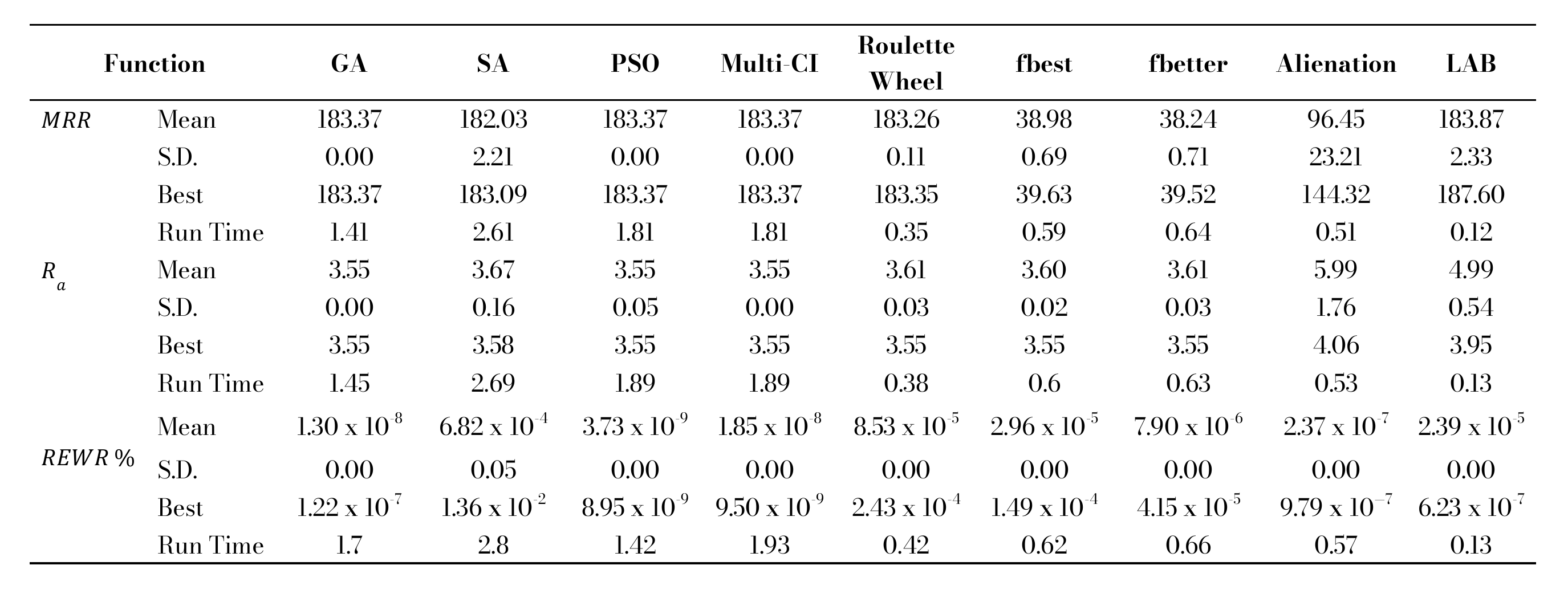}
\label{tab:CompareEDM1}
\captionsetup{singlelinecheck = false, format= hang, justification=raggedright, font=footnotesize, labelsep=space}
\caption{Overall solutions to $R_a$, $MRR$, $REWR$ of EDM}
\includegraphics[width=0.95\linewidth]{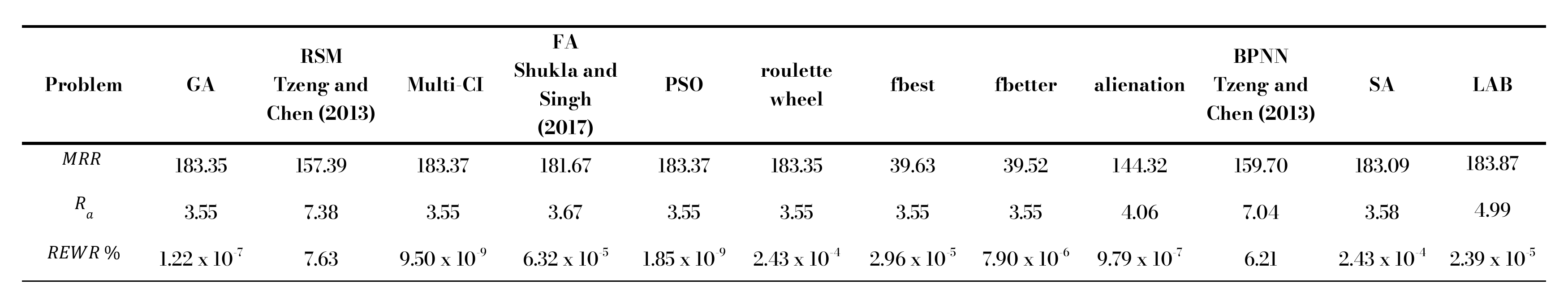}
\centering
\label{tab:CompareEDM2}
\end{table}
\end{landscape}

\subsection{Solutions to Micro-machining problems}
Comparison Tables \ref{tab:Turning}, \ref{tab:Milling} and \ref{tab:Drilling} exhibit  
solutions consisting of mean and best solution along with standard deviation for 30 trials of each objective function of algorithms for solving micro-turning, micro-milling and micro-drilling processes.
For micro-drilling processes, LAB obtained comparable results with Mulit-CI, GA, SA and variations of CI as well as outperforming GA, SA and PSO in convergence rate. However, for micro-turning and micro-milling with $0.7 mm$ and with $1 mm$ tool diameter for machining time ($M_t$) LAB could compete with other algorithms but could not produce superior results.

\begin{table}[H]
\captionsetup{singlelinecheck = false, format= hang, justification=raggedright, font=footnotesize, labelsep=space}
\caption{Solutions to Micro-Turning processes}
\includegraphics[width=0.95\linewidth]{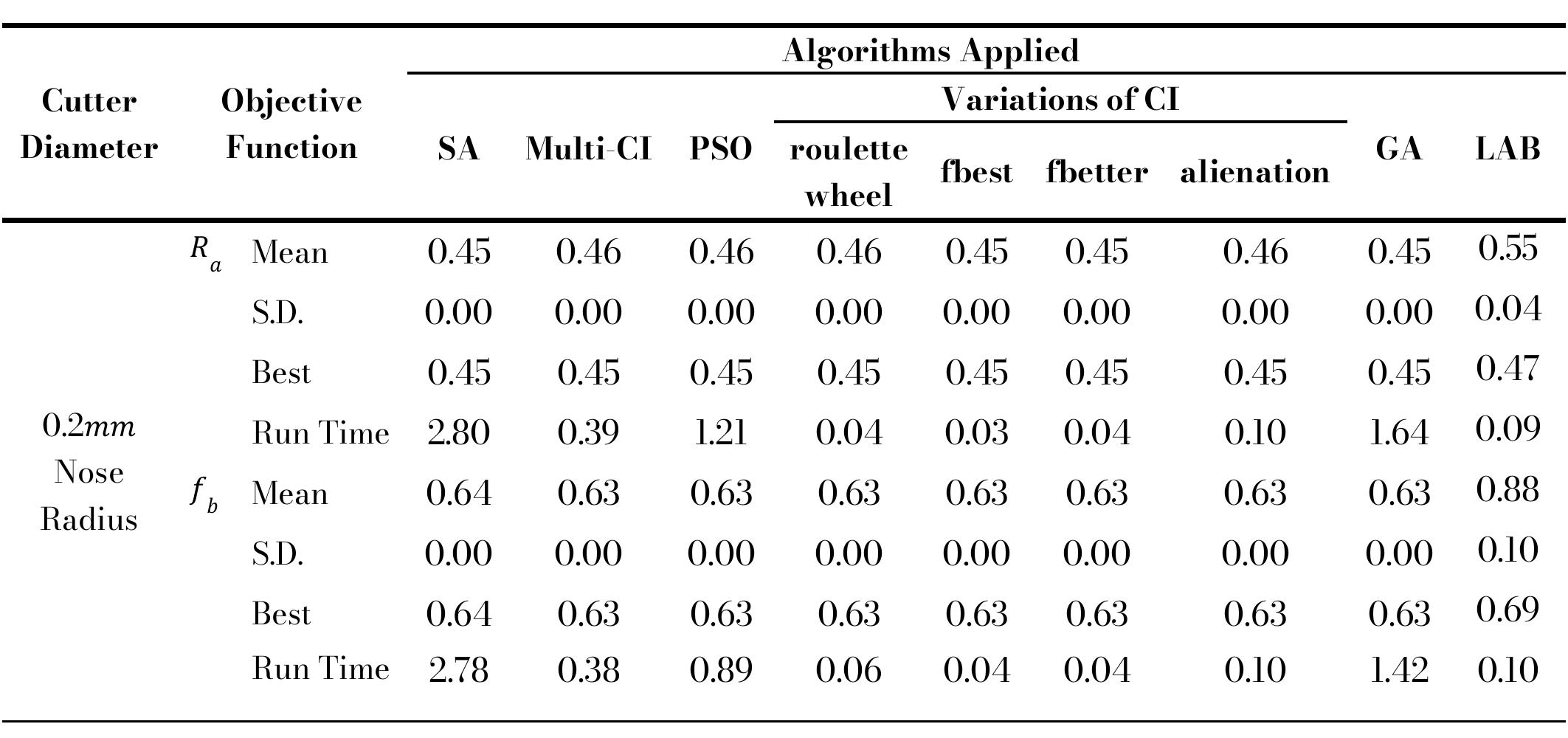}
\centering
\label{tab:Turning}
\end{table}

\begin{figure}[H]
\centering
\begin{subfigure}{0.99\textwidth}
    \includegraphics[width=\textwidth]{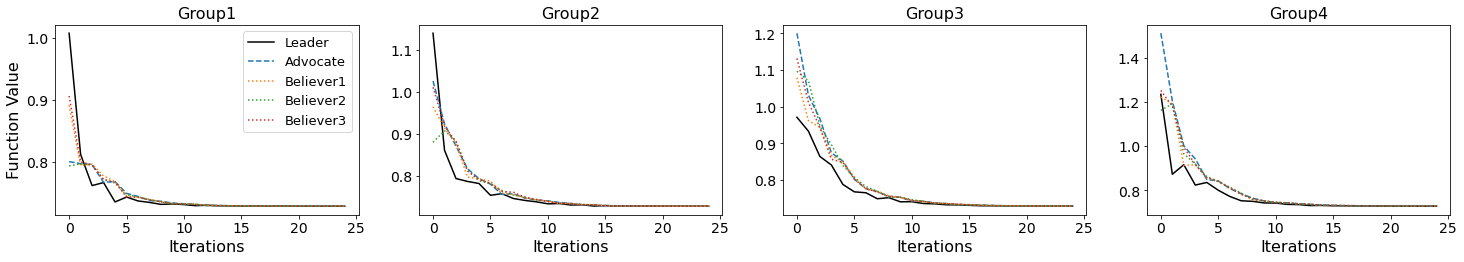}
    \caption{$f_b$}
    \label{fig:fb}
\end{subfigure}
\begin{subfigure}{0.99\textwidth}
    \includegraphics[width=\textwidth]{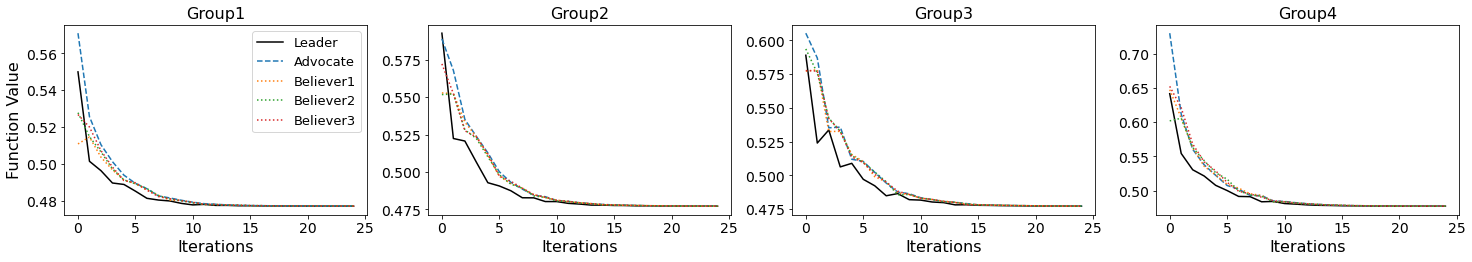}
    \caption{$R_a$}
    \label{fig:Ra}
\end{subfigure}
\caption{Convergence: Micro-Turning}
\label{fig:micro-turning}
\end{figure}

\begin{table}[!htb]
\captionsetup{singlelinecheck = false, format= hang, justification=raggedright, font=footnotesize, labelsep=space}
\caption{Solutions to Micro-Milling processes}
\includegraphics[width=0.95\linewidth]{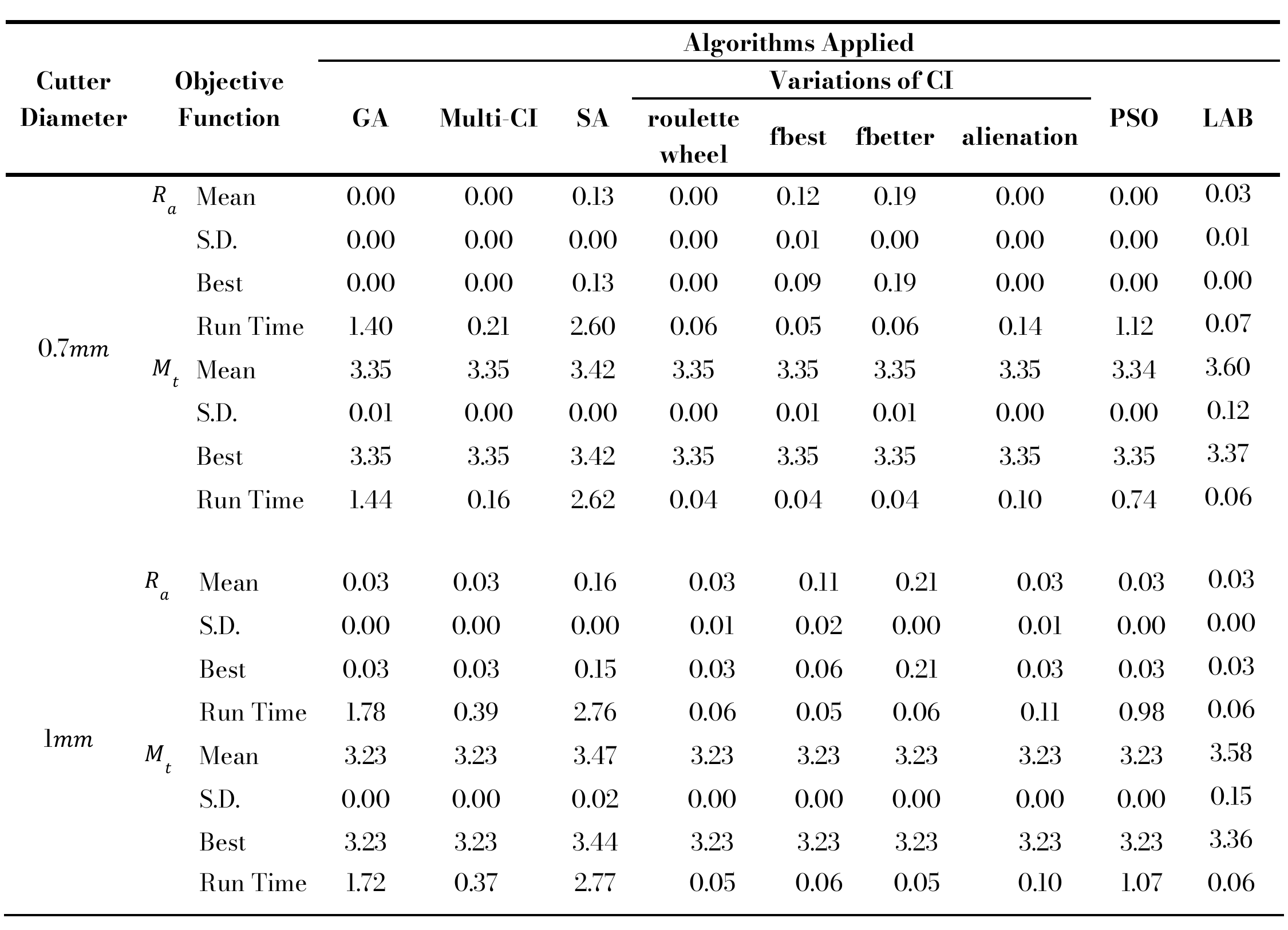}
\centering
\label{tab:Milling}
\end{table}

\begin{figure}[!htb]
\centering
\begin{subfigure}{0.99\textwidth}
    \includegraphics[width=\textwidth]{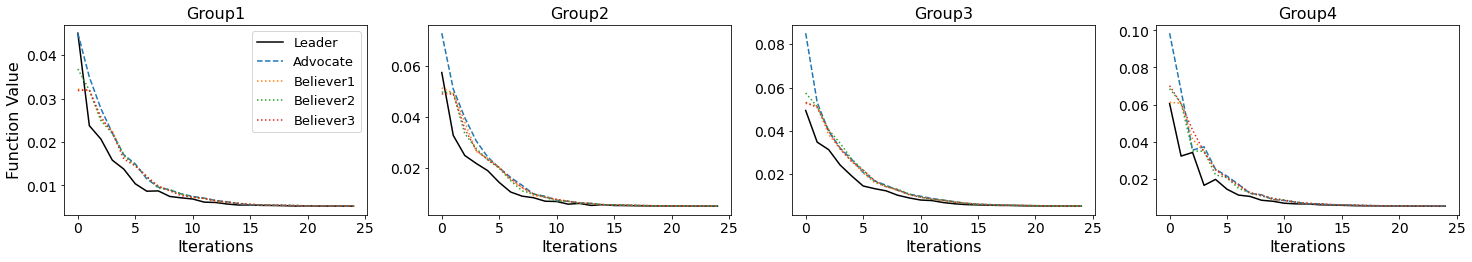}
    \caption{$R_a(0.7mm)$}
    \label{fig:Ra0.7}
\end{subfigure}
\begin{subfigure}{0.99\textwidth}
    \includegraphics[width=\textwidth]{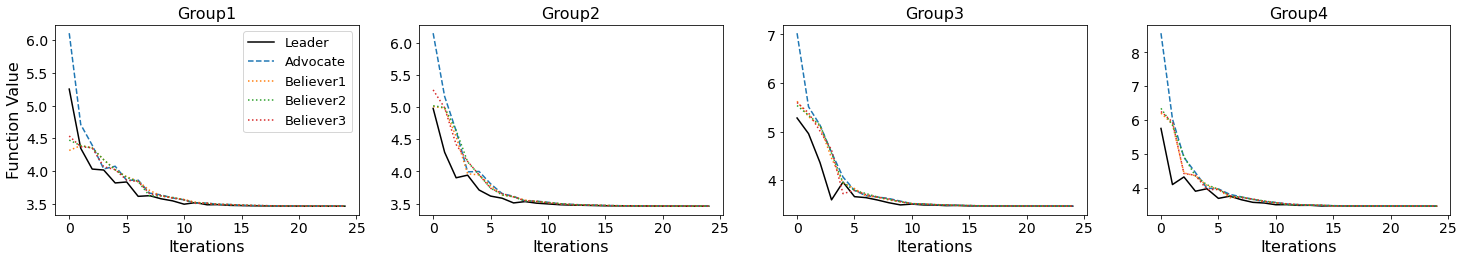}
    \caption{$M_t(0.7mm)$}
    \label{fig:Mt0.7}
\end{subfigure}
\begin{subfigure}{0.99\textwidth}
    \includegraphics[width=\textwidth]{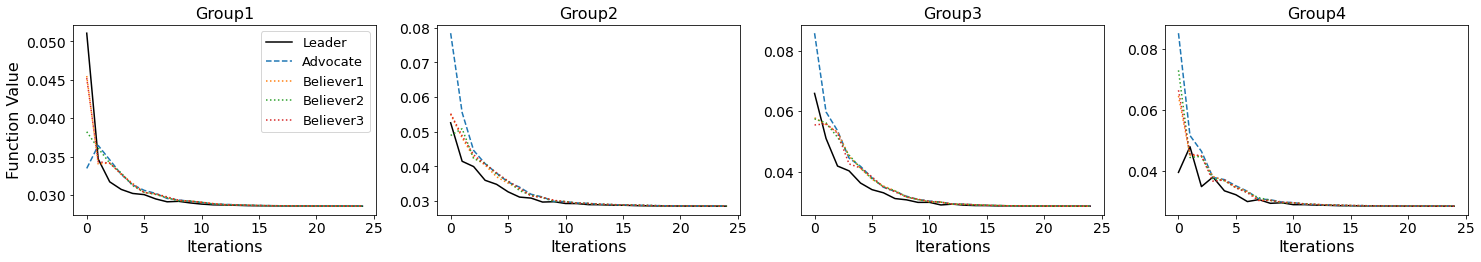}
    \caption{$R_a(1mm)$}
    \label{fig:Ra1}
\end{subfigure}
\begin{subfigure}{0.99\textwidth}
    \includegraphics[width=\textwidth]{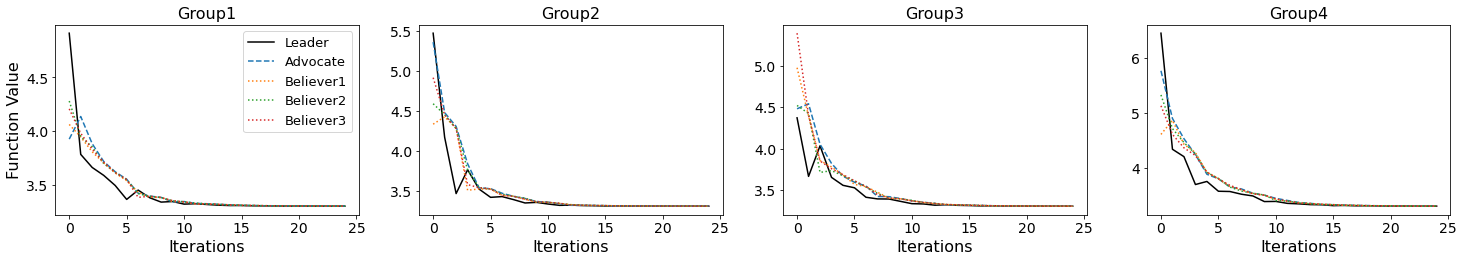}
    \caption{$M_t(1mm)$}
    \label{fig:Mt1}
\end{subfigure}
\caption{Convergence: Micro-Milling}
\label{fig:micro-milling}
\end{figure}

\begin{table}[!htb]
\captionsetup{singlelinecheck = false, format= hang, justification=raggedright, font=footnotesize, labelsep=space}
\caption{Solutions to Micro-Drilling processes}
\includegraphics[width=0.95\linewidth]{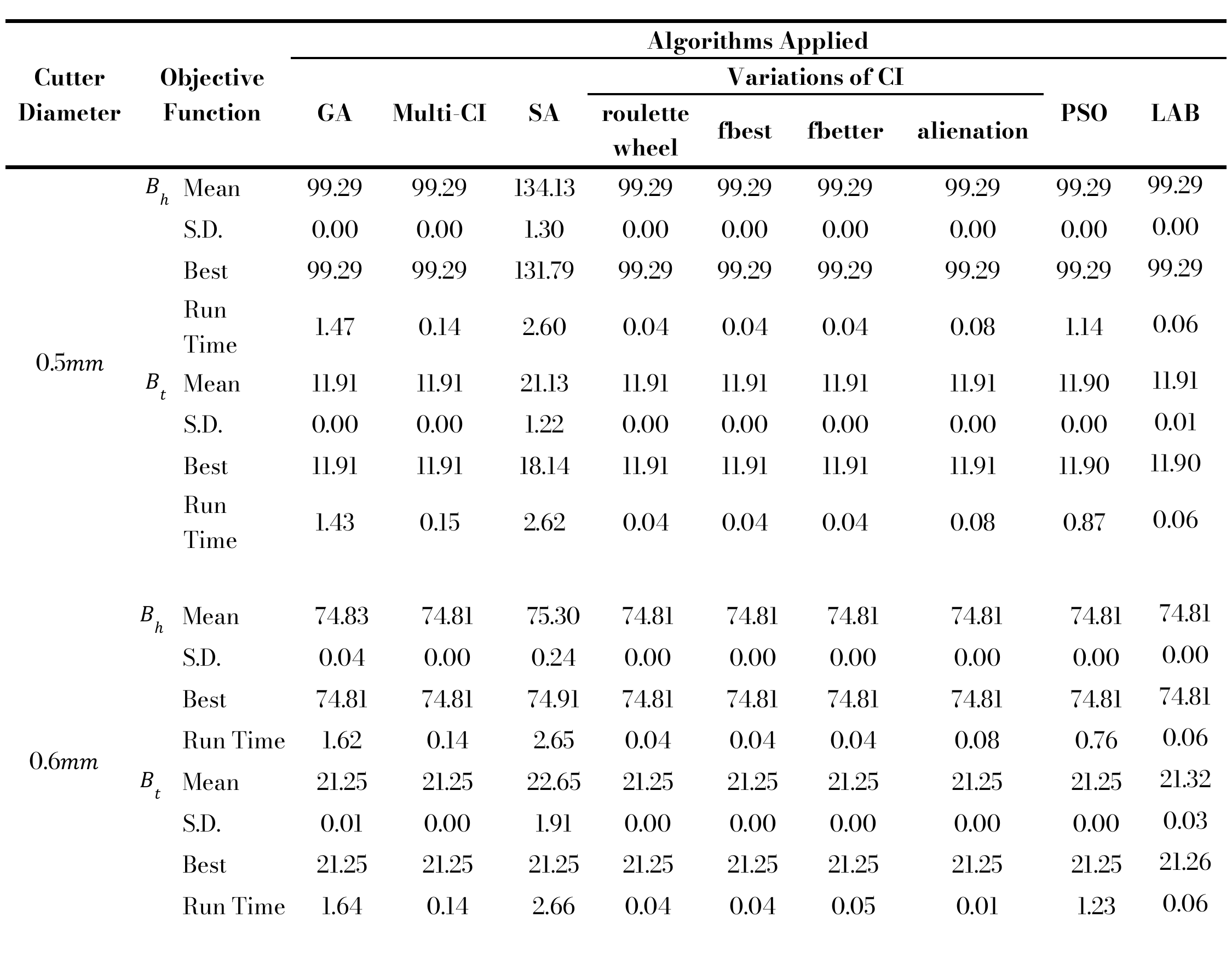}
\centering
\label{tab:Drilling}
\end{table}

\pagebreak

\begin{table}[!htb]
\captionsetup{singlelinecheck = false, format= hang, justification=raggedright, font=footnotesize, labelsep=space}
\caption*{\textbf{Table \ref{tab:Drilling} } Continued}
\includegraphics[width=0.95\linewidth]{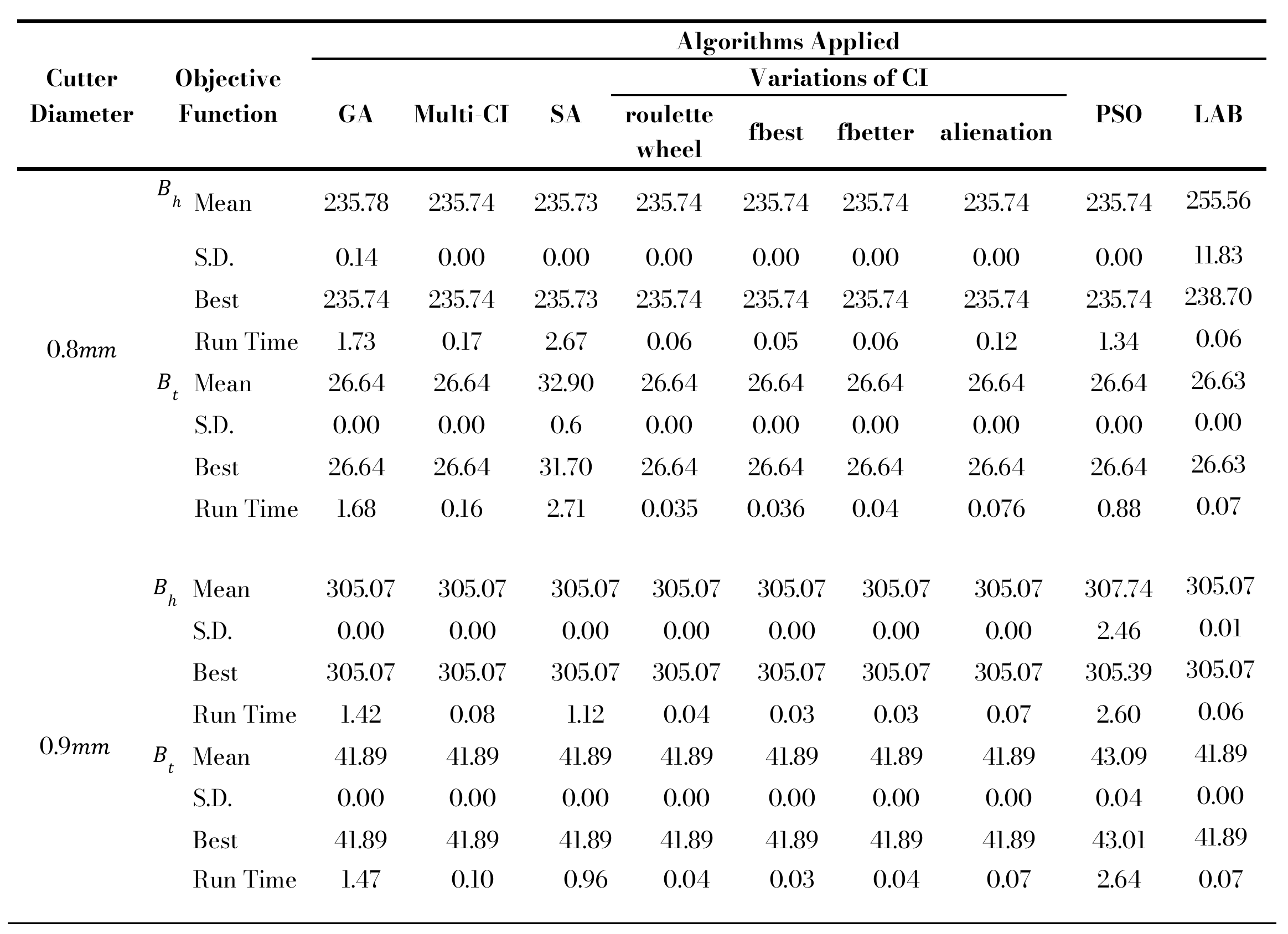}
\centering
\end{table}

\begin{figure}[!htb]
\centering
\begin{subfigure}{0.99\textwidth}
    \includegraphics[width=\textwidth]{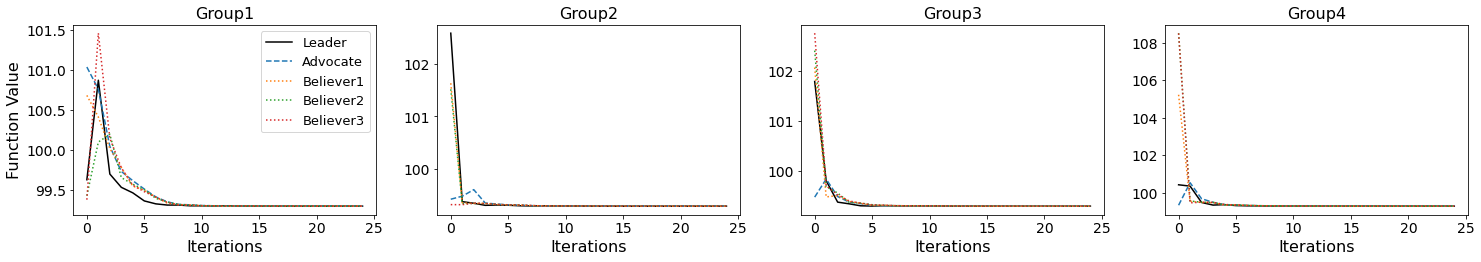}
    \caption{$B_h(0.5mm)$}
    \label{fig:Bh0.5}
\end{subfigure}
\begin{subfigure}{0.99\textwidth}
    \includegraphics[width=\textwidth]{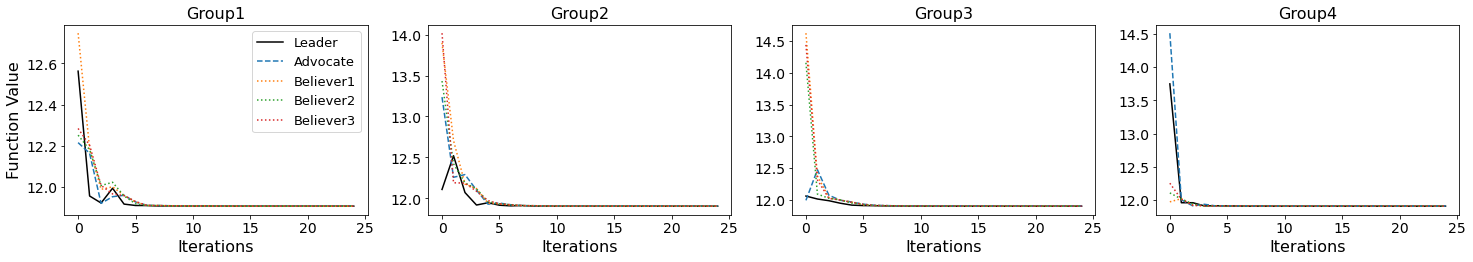}
    \caption{$B_t(0.5mm)$}
    \label{fig:Bt0.5}
\end{subfigure}
\begin{subfigure}{0.99\textwidth}
    \includegraphics[width=\textwidth]{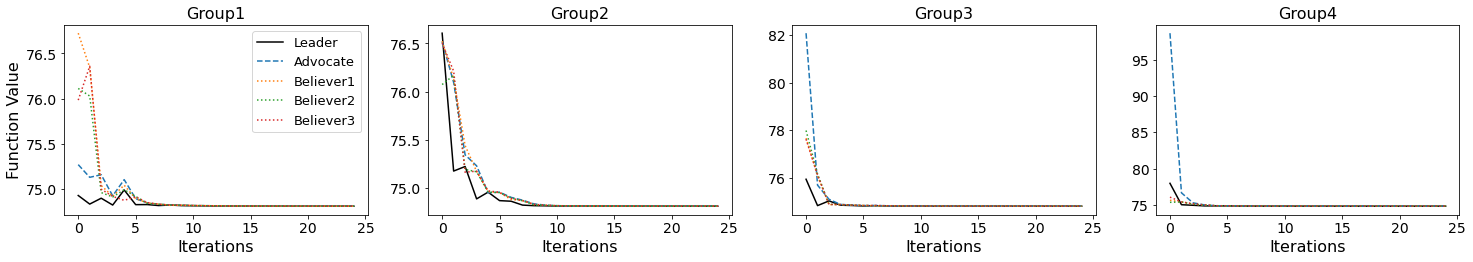}
    \caption{$B_h(0.6mm)$}
    \label{fig:Bh0.6}
\end{subfigure}
\begin{subfigure}{0.99\textwidth}
    \includegraphics[width=\textwidth]{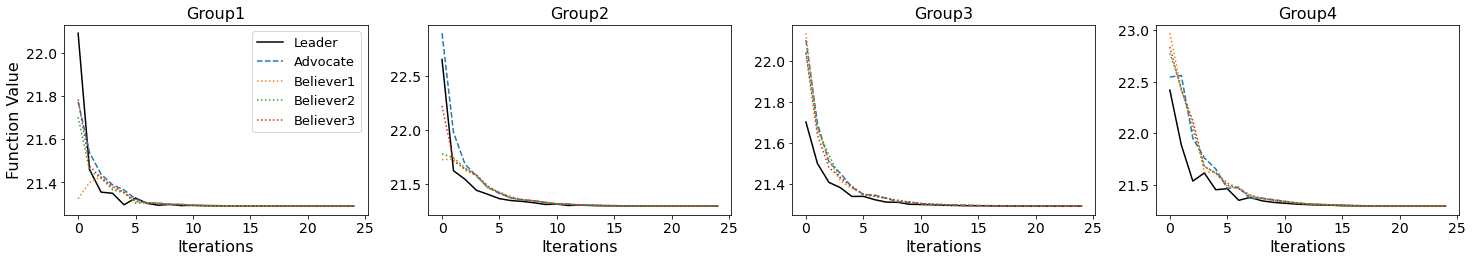}
    \caption{$B_t(0.6mm)$}
    \label{fig:Bt0.6}
\end{subfigure}
\label{fig:micro-drilling}
\end{figure}

\FloatBarrier
\begin{figure}[!htb]
\centering
\ContinuedFloat
\begin{subfigure}{0.99\textwidth}
    \includegraphics[width=\textwidth]{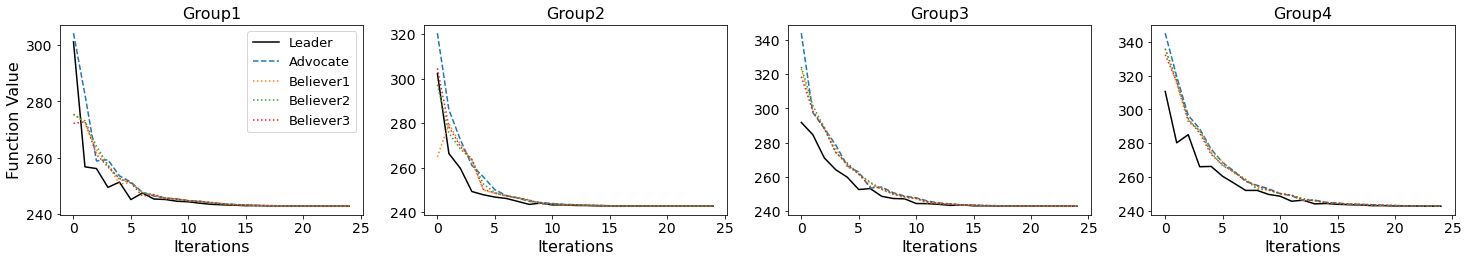}
    \caption{$B_h(0.8mm)$}
    \label{fig:Bh0.8}
\end{subfigure}
\begin{subfigure}{0.99\textwidth}
    \includegraphics[width=\textwidth]{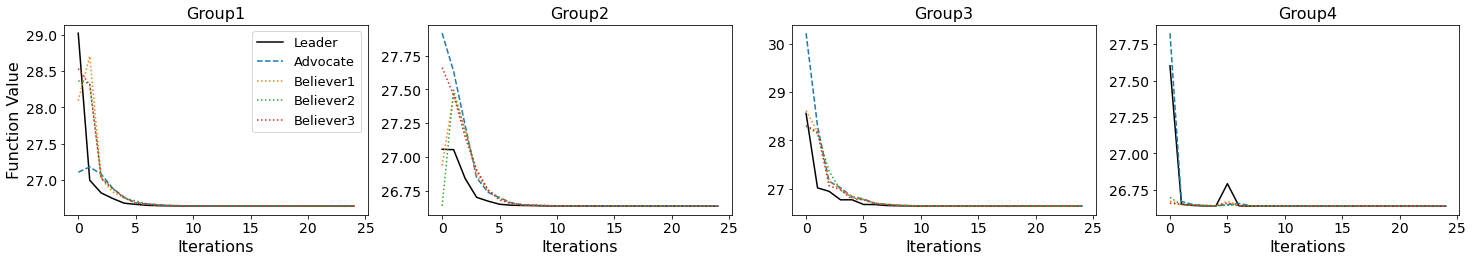}
    \caption{$B_t(0.8mm)$}
    \label{fig:Bt0.8}
\end{subfigure}
\begin{subfigure}{0.99\textwidth}
    \includegraphics[width=\textwidth]{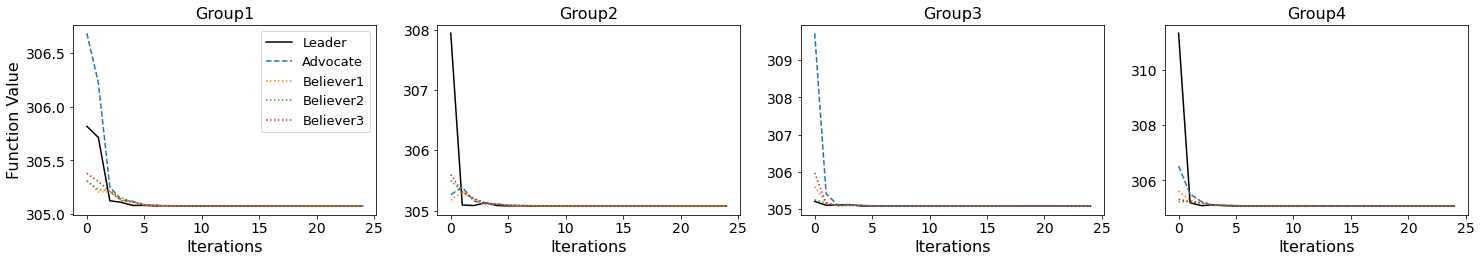}
    \caption{$B_h(0.9mm)$}
    \label{fig:Bh0.9}
\end{subfigure}
\begin{subfigure}{0.99\textwidth}
    \includegraphics[width=\textwidth]{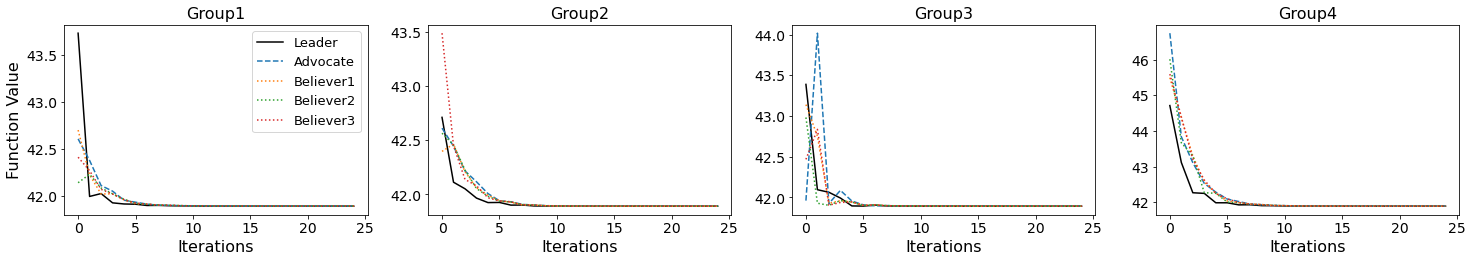}
    \caption{$B_t(0.9mm)$}
    \label{fig:Bt0.9}
\end{subfigure}
\caption{Convergence: Micro-Drilling}
\label{fig:micro-drilling2}
\end{figure}

The LAB solutions exhibited higher standard deviation for micro-turning (Table \ref{tab:Turning}), micro-drilling (Table \ref{tab:Milling}) and micro-milling (Table \ref{tab:Drilling}) problems, for convergence plots refer to Fig. \ref{fig:micro-turning}, \ref{fig:micro-milling} and \ref{fig:micro-drilling2}, respectively. This is because the individuals in LAB are updated at every iteration after computing individual search directions, to simultaneously obtain updated solutions and rankings. This iterative individual updating process after computing individual search direction i.e local ranking as well as global ranking, thus resulting less robustness and higher standard deviation. 

When comapred with other algorithms for solving micro-machining problems LAB resulted in lower run time as compared to other algorithms but showed less robustness. However, LAB outperformed SA, $f_{best}$ and $f_{better}$ by achieving 76\%, 85\% and 75\% minimization of $R_a$ respectively for micro-milling with 0.7 mm tool diameter. LAB achieved 81\%, 72\%, 85\% minimization of $R_a$ when compared to SA, $f_{best}$ and $f_{better}$ for 1 mm tool diameter. LAB also achieved 24\% and 34\% minimization of $B_h$ and $B_t$ as compared to SA for micro-drilling with tool diameter 0.5 mm. For tool diameter 0.8 mm and 0.9 mm, 16\% and 3\% minimization of $B_t$, respectively, were achieved as compared to SA (exhibited in Tables \ref{tab:Turning}, \ref{tab:Milling}, \ref{tab:Drilling}).

\subsection{Solution to Turning of Titanium Alloy}

\begin{figure}[!htb]
\centering
\begin{subfigure}{0.99\textwidth}
    \includegraphics[width=\textwidth]{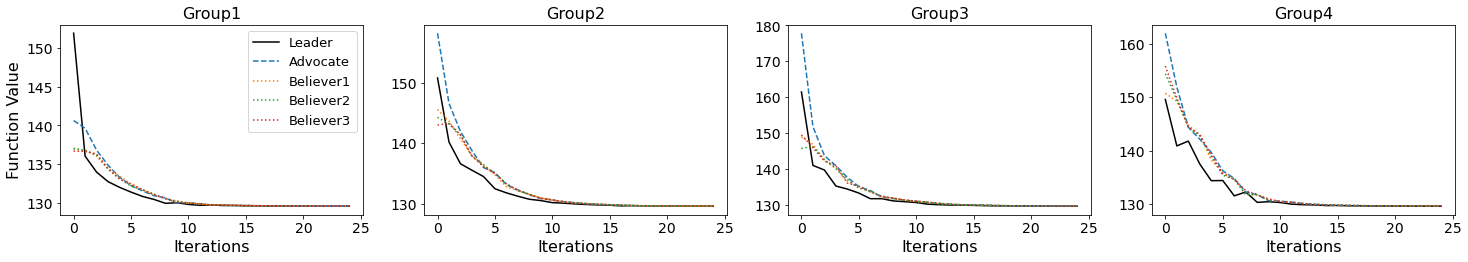}
    \caption{$F_c$}
    \label{fig:SN_Fc}
\end{subfigure}
\begin{subfigure}{0.99\textwidth}
    \includegraphics[width=\textwidth]{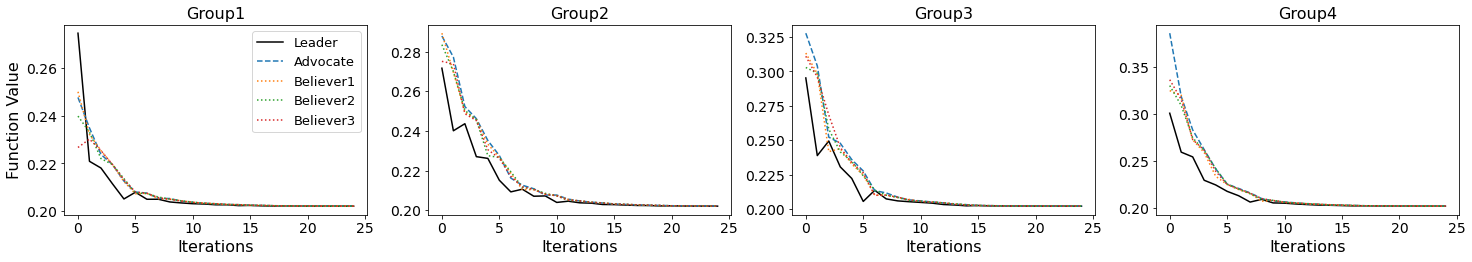}
    \caption{$V_{Bmax}$}
    \label{fig:VBmax}
\end{subfigure}
\begin{subfigure}{0.99\textwidth}
    \includegraphics[width=\textwidth]{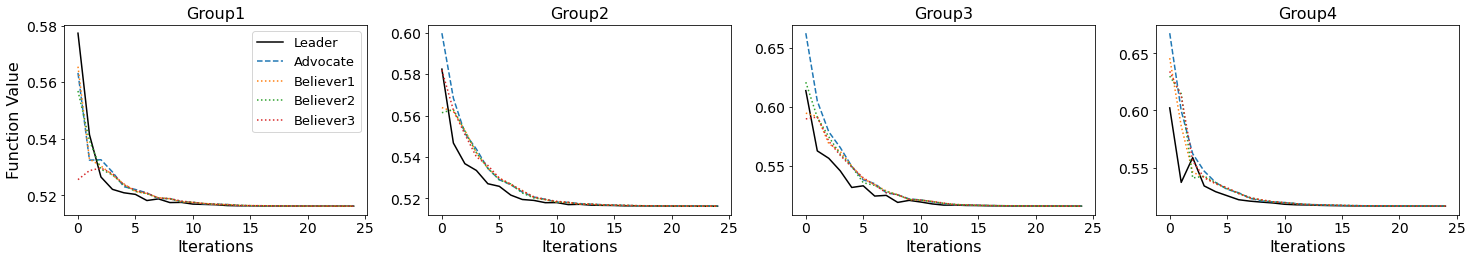}
    \caption{$L$}
    \label{fig:L}
\end{subfigure}
\begin{subfigure}{0.99\textwidth}
    \includegraphics[width=\textwidth]{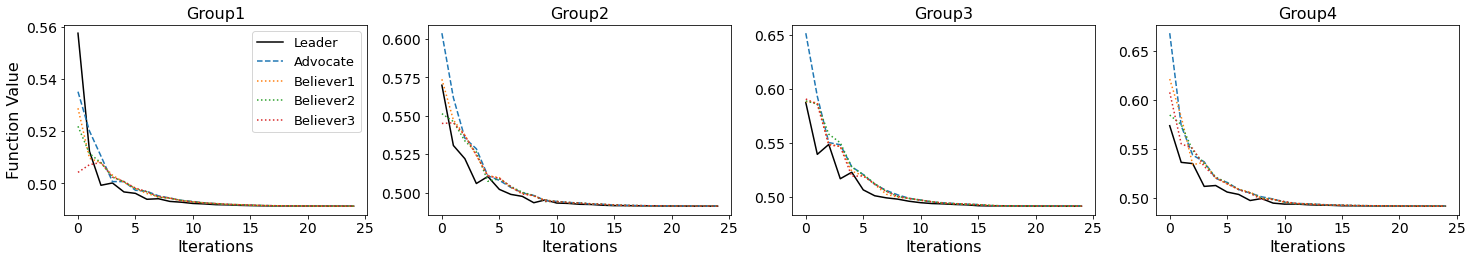}
    \caption{$R_a$}
    \label{fig:SNRa}
\end{subfigure}
\caption{Convergence Plots for optimal values of $F_c, V_{Bmax}, L, R_a$}
\label{fig:SN}
\end{figure}

Table \ref{tab:SN1} includes best solutions obtained for Cutting Force $F_c$, Tool Wear $V_{Bmax}$, Tool Chip Contact Length $L$ and Surface Roughness $R_a$ produced by variations of CI, Multi-CI and LAB with their corresponding mean solutions, standard deviation and run time.
Table \ref{tab:SN2} contains additional comparison of solutions by algorithms namely experimental work, desirability approach and PSO. In Table \ref{tab:SN3} optimum values yielded by variations of CI, Multi-CI and LAB for cutting speed $V_c$ , feed $f$ and the tool angle $\phi$ are shown. The algorithm needs more balanced exploration and exploitation abilities to find global optimum solution as it is quite evident from Eq. \ref{MQL_Fc} it is inseparable, multimodal and nonlinear in nature. Plots in Fig.\ref{fig:SN_Fc}, \ref{fig:VBmax}, \ref{fig:L} and \ref{fig:SNRa} represent the best solutions of LAB for cutting force $F_c$ , tool wear $V_{Bmax}$, tool-chip contact length $L$ and surface roughness $R_a$ respectively. Efforts of the individuals in climbing up the rankings by competing to be the best are evident in Fig. \ref{fig:SN}.

\begin{table}[!htb]
\captionsetup{singlelinecheck = false, format= hang, justification=raggedright, font=footnotesize, labelsep=space}
\caption{Comparison of statistical solutions for Turning in MQL environment}
\includegraphics[width=0.95\linewidth]{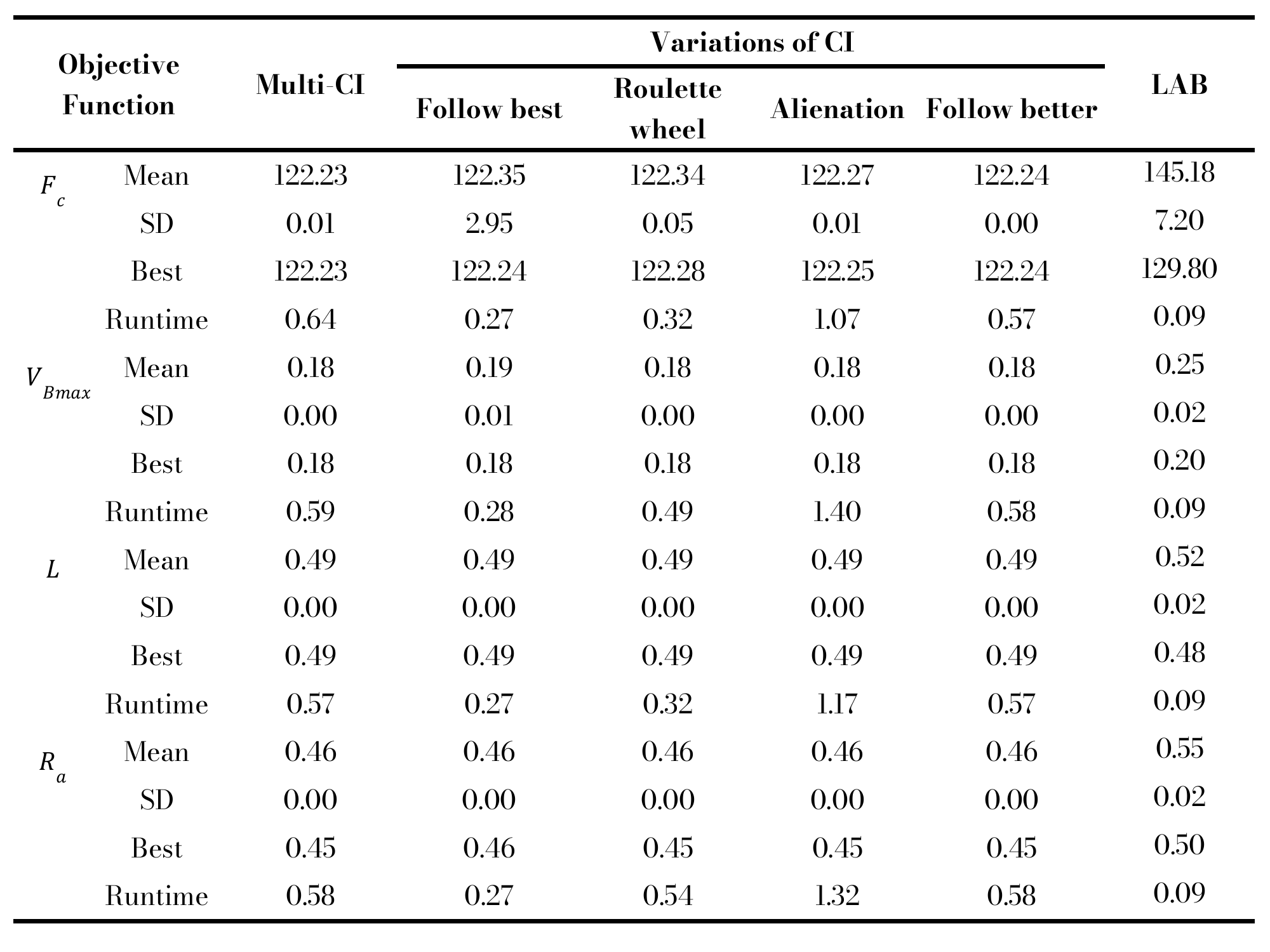}
\centering
\label{tab:SN1}
\end{table}

\begin{table}[!htb]
\captionsetup{singlelinecheck = false, format= hang, justification=raggedright, font=footnotesize, labelsep=space}
\caption{Comparison of algorithms}
\includegraphics[width=0.95\linewidth]{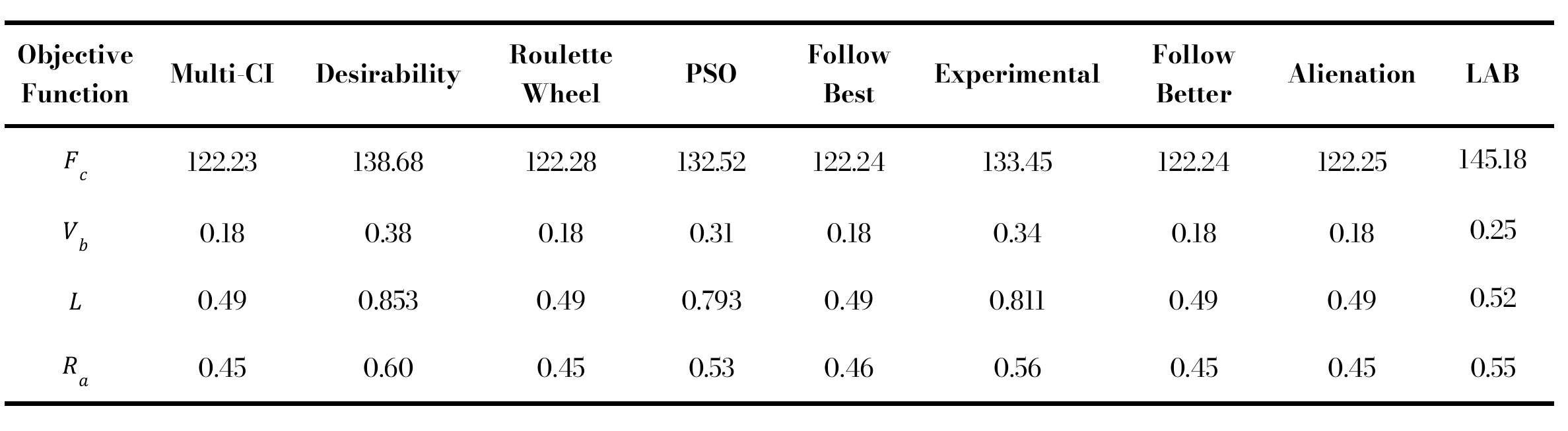}
\centering
\label{tab:SN2}
\end{table}

\begin{table}[!htb]
\captionsetup{singlelinecheck = false, format= hang, justification=raggedright, font=footnotesize, labelsep=space}
\caption{Comparison of optimum values for the solutions of $V_c$, $f$, and $\phi$}
\includegraphics[width=0.95\linewidth]{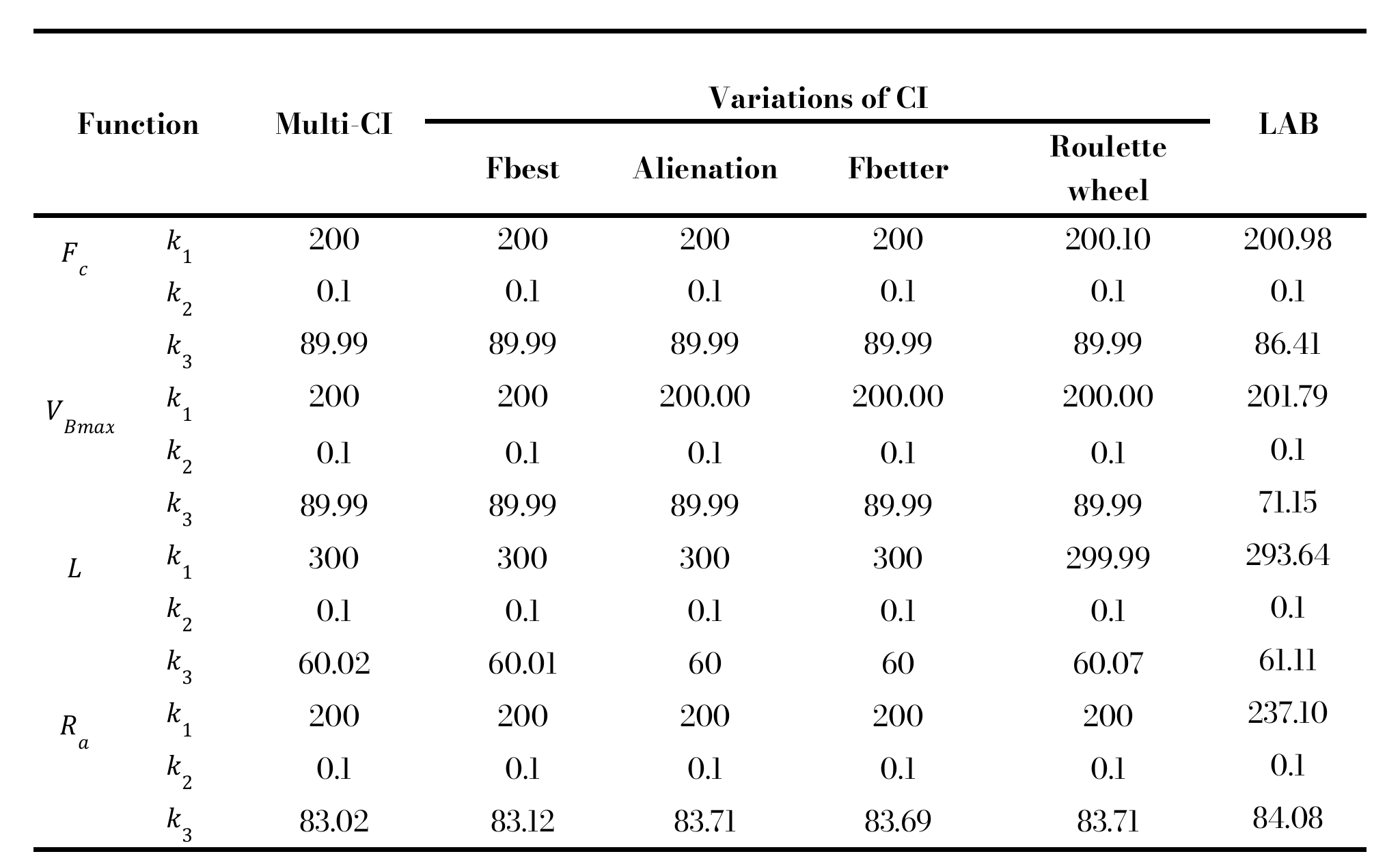}
\centering
\label{tab:SN3}
\end{table}

\FloatBarrier

\section{Conclusions and future directions}
In this manuscript, a novel socio-inspired algorithm is introduced, named the LAB algorithm, based on how individuals in a group with certain personality traits follow, make decisions and compete within the group in society. The proposed algorithm was examined by solving 27 benchmark test problems from CEC 2005 and a statistical comparison using Wilcoxon-signed rank test was conducted. LAB was able to perform slightly better when compared in terms of best solution, mean solution, robustness and computational time when compared to CMAES and IA and was able to outperform PSO2011, CMAES, ABC, JDE, CLPSO, and SADE in computational time. LAB demonstrated low robustness but exceedinly low computational time.

The algorithm was also validated by solving 23 real-world problems consisting of AWJM, EDM, Parameter tuning of turning titanium alloy and Advanced manufacturing processes problem to compare exploitation, exploration, computation cost and convergence rate with other well-known and recent algorithms: Experimental (Kechigas, 2012), Regression (Kechigas, 2012), FA, Variations of CI (roulette wheel, $f_{best}$, $f_{better}$, alienation), GA, SA, PSO, Multi-CI.

Problems for minimization of surface roughness $R_a$ for AWJM, EDM and micro-machining processes namely micro-turning and micro-milling for Advanced Manufacturing Processes were solved.
Minimization of burr thickness $B_t$ and burr height $B_h$, relative electrode wear rate $REWR$ for EDM and taper angle $kerf$ for AWJM in micro-drilling was executed.
In micro-turning process flank wear $f_b$ and in micro-milling processes machining time $M_t$ were minimized.
Micro-drilling process utilized four drilling cutter diameters: $0.5 mm$; $0.6 mm$; $0.8 mm$ and $0.9 mm$. In the micro-milling processes, two cutter diameters: $0.7 mm$ and $1 mm$, were utilized. The results of LAB were then compared with mutltiple algorithms consisting of variations of CI, Multi-CI algorithm, experimental results and also with relatively modern algorithms such as SA, PSO, GA, BPNN, RSM and FA.

LAB was able to perform exceedingly well when compared to FA, SA, PSO, experimental results and solutions using regression for solving $kerf$ of AWJM problem in terms of solution quality. LAB results were comparable with GA and PSO for solving EDM and micro-machining problems. LAB was able to outperform variations of CI, regression, RSM, FA, SA, BPNN approaches in terms of solutions obtained. The run time of LAB is quite lower as compared to other algorithms for majority of the problems, because in LAB all the individuals simultaneously compete and interact with one another and individuals are updated at every iteration helps it gain more exploration and exploitation capabilities; however, it resulted in higher standard deviation which exhibited its low robustness.

Several enhancements can be done in the algorithm for better and faster computation in order to solve complex and higher dimension problems easily, by introducing a method of triggering the algorithm when stuck at local minima, which may help LAB solve a wider range of higher dimension complex real-life problems. Moreover, LAB algorithm can be modified to solve multi-objective problems making the competitive groups to handle different objectives.
\pagebreak

\begin{large}
    \textbf{Acknowledgments}
\end{large} 
This work was supported by the Fundamental Research Grant Scheme (FRGS) under the Ministry of Higher Education (MOHE) with project number FRGS/1/2020/ICT02/ MUSM/03/6

\printbibliography
\end{document}